\documentclass[11pt]{article}

\usepackage[preprint]{acl}

\usepackage{times}
\usepackage{latexsym}

\usepackage[T1]{fontenc}

\usepackage[utf8]{inputenc}

\usepackage{microtype}

\usepackage{inconsolata}

\usepackage{graphicx}

\usepackage{hyperref}
\usepackage{url}

\usepackage{subfig}
\usepackage{amsmath}
\usepackage{amssymb}

\usepackage{algorithm}
\usepackage{algorithmic}

\usepackage{stfloats} 

\usepackage{booktabs}
\usepackage{multirow}
\usepackage{colortbl}
\usepackage[normalem]{ulem}

\usepackage{arydshln}
\usepackage{xcolor}
\definecolor{darkred}{RGB}{200, 0, 0}
\definecolor{darkblue2}{RGB}{49,119,183}
\definecolor{lightgreen}{RGB}{229, 233, 218}
\definecolor{darkgreen}{RGB}{0, 100, 0}
\definecolor{lightpurple}{RGB}{222,217,231}
\definecolor{darkpurple}{RGB}{119,102,155}
\definecolor{lightyellow}{RGB}{250,250,231}
\definecolor{lightblue}{RGB}{221,235,250}
\definecolor{lightorange}{RGB}{250,237,220}
\definecolor{bluegray}{RGB}{217,227,232}
\usepackage{soul}
\sethlcolor{lightblue}

\usepackage{bxcoloremoji/bxcoloremoji}

\usepackage{enumitem}
\usepackage{dsfont}

\usepackage{wrapfig}

\usepackage[misc]{ifsym}
\usepackage{footmisc}
\DefineFNsymbols*{newfootnote}{%
    \textasteriskcentered{\textrm{\Letter}}\dag\textdaggerdbl{\ding{73}}\P{**}%
    {\ding{172}}{\ding{173}}{\ding{174}}}
\setfnsymbol{newfootnote}

\title{\textsc{\model}: Towards Agentic Tool-Use Reward Modeling}

\newcommand*\samethanks[1][\value{footnote}]{\footnotemark[#1]}
\author{
Renhao Li$^{1,4}$\thanks{Work done during an internship at Qwen Team.},\quad
Jianhong Tu$^2$,\quad
Yang Su$^2$,\quad
Yantao Liu$^2$,\quad
Fei Huang$^2$,\\
\textbf{Hamid Alinejad-Rokny$^3$,}\quad
\textbf{Derek F. Wong$^{1}$\thanks{Corresponding author.},}\quad
\textbf{Junyang Lin$^{2}$\samethanks,}\quad
\textbf{Min Yang$^{4}$\samethanks}\\[0.5em]
$^1$University of Macau,
$^2$Qwen Team, Alibaba Inc., 
$^3$UNSW Sydney\\
$^4$Shenzhen Institute of Advanced Technology, Chinese Academy of Sciences\\
\texttt{li.renhao@connect.um.edu.mo,
derekfw@um.edu.mo}\\
\texttt{junyang.ljy@alibaba-inc.com,
min.yang@siat.ac.cn}\\[0.5em]
\href{https://huggingface.co/collections/RioLee/toolrm}{\raisebox{-0.2em}{\includegraphics[height=1.1em]{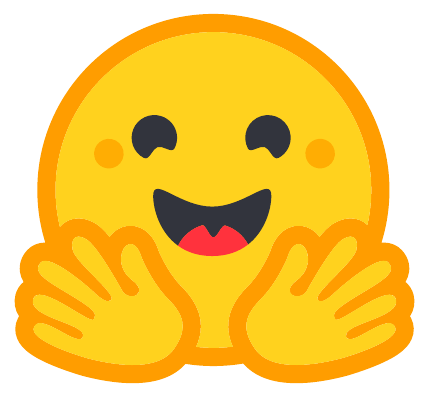}} \texttt{Dataset}}
\quad\quad
\href{https://github.com/lirenhao1997/ToolRM}{\raisebox{-0.2em}{\includegraphics[height=1.1em]{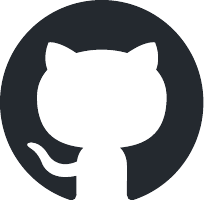}} \texttt{Code}}\\[0.5em]
}

\newcommand{\model}{ToolRM}
\newcommand{\benchmark}{TRBench$_{\text{BFCL}}$}
\newcommand{\dataset}{ToolPref-Pairwise-30K}

\begin{document}
\maketitle


\begin{abstract}
Reward models (RMs) play a critical role in aligning large language models (LLMs) with human preferences. Yet in the domain of tool learning, the lack of RMs specifically designed for function-calling tasks has limited progress toward more capable agentic AI. We introduce \textsc{\model}, a family of lightweight reward models tailored for general tool-use scenarios. To build these models, we propose a novel pipeline that constructs high-quality pairwise preference data using rule-based scoring and multidimensional sampling. This yields \textit{\dataset}, a diverse, balanced, and challenging preference dataset that supports both generative and discriminative reward modeling. We also introduce \textsc{\benchmark}, a benchmark built on the agent evaluation suite BFCL to evaluate RMs on tool calling tasks. Trained on our constructed data, models from the Qwen3-4B/8B series achieve up to 17.94\% higher accuracy, substantially outperforming frontier LLMs and RMs in pairwise reward judgments. Beyond training objectives, generative {\model} generalizes to broader critique tasks, including Best-of-N sampling and self-correction. Experiments on ACEBench highlight its effectiveness and efficiency, enabling inference-time scaling while reducing output token usage by over 66\%. Its support for downstream RL training further validates its practical utility.
We release data to facilitate future research.
\end{abstract}

\section{Introduction}
\label{sec:intro}
Recent advances in agentic artificial intelligence have been driven in large part by the tool-use capabilities~\citep{patil2024gorilla,openai2025deepresearch} of large language models (LLMs). By leveraging external tools, LLMs can recognize their limitations and extend their capabilities through environment interaction. The research focus has recently shifted from behavior cloning via supervised finetuning on curated trajectories~\citep{schick2023toolformer,tang2023toolalpaca} to trial-and-error approaches based on reinforcement learning from verifiable rewards~\cite[RLVR,][]{feng2025retool,qian2025toolrl}, enabling generalizable and robust tool-use behavior.

Despite these gains, the lack of reliable reward models (RMs) tailored to tool-use tasks remains a core limitation. Most existing methods depend on verified tool-call trajectories for feedback, which restricts scalability to domains lacking such annotations. At inference time, the absence of precise reward signals also makes it hard to leverage multiple sampled answers for test-time selection~\citep{wang2023selfconsistency,snell2025scaling}. We argue that developing a robust RM—capable of evaluating tool-use behavior without requiring ground-truth labels—is critical for advancing this field.
Building effective RMs for tool-use presents three key challenges: 
\textbf{(C1)} Constructing high-quality preference pairs that reflect tool-use intent. 
\textbf{(C2)} Enabling generalizable critique beyond 3H-style modeling~\citep{askell2021general}, as tool-use tasks often allow more objective, causal reasoning. 
\textbf{(C3)} Evaluating RM performance in this setting, which remains underexplored for frontier LLMs and specialized critics.

\begin{figure*}[!htb]
\centering
\includegraphics[width=0.98\linewidth]{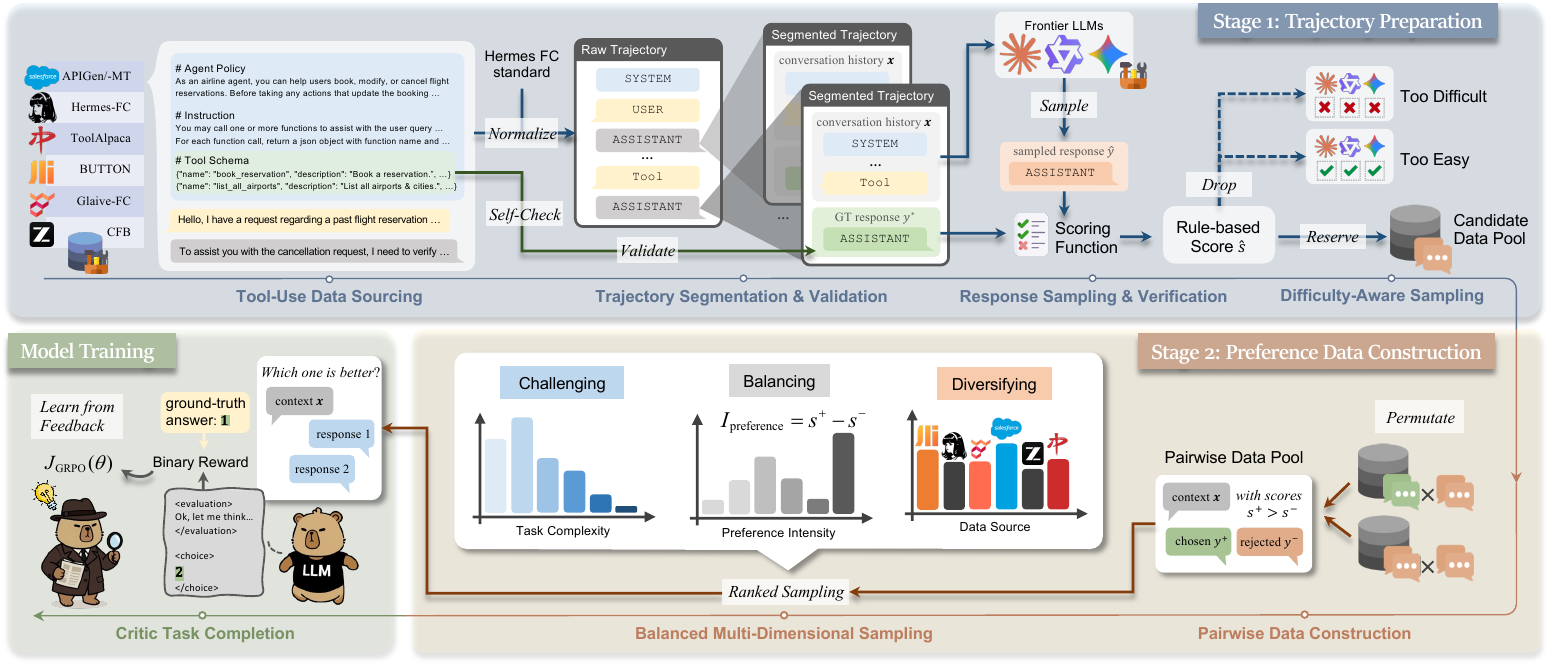}
\caption{Overview of the proposed pipeline for training {\model}.}
\label{fig:framework}
\end{figure*}

To address these challenges, we introduce \textsc{\model}, a family of lightweight RMs for general tool-use tasks, along with a two-stage pipeline for constructing high-quality preference data to train them. First, we curate and validate tool-calling trajectories from diverse open-source datasets, segment them into context–response pairs, and sample alternative responses using multiple LLMs. Instead of relying on ground-truth matches, we apply rule-based labeling to capture fine-grained preferences. A multidimensional sampling strategy ensures diverse scenarios, varied preference intensity, and high task complexity \textbf{(C1)}. To strengthen critique ability, we train generative {\model} with a pairwise objective using unified instructions and verifiable supervision, enabling the model to learn robust reasoning without curated traces \textbf{(C2)}. We also introduce \textsc{\benchmark}, a benchmark based on BFCL~\citep{patil2025bfcl}, to systematically evaluate RM performance on tool-use tasks \textbf{(C3)}.

Our contributions are summarized as follows:
\begin{itemize}[leftmargin=*]
    \item We propose a novel pipeline for generating high-quality pairwise preference data for reward modeling in tool-use scenarios. Using seven open-source tool-calling datasets, we construct \textit{\dataset}, a diverse and balanced set of 30,000 challenging preference pairs. This resource is publicly released to support future work in tool-oriented reward modeling.
    
    \item We train \textsc{\model} on the Qwen3 families with different objectives, achieving substantial gains in pairwise reward judgments. The resulting models generalize to broader critique tasks, enabling efficient inference-time scaling, producing compact, high-quality critiques, and supporting effective downstream RL training.
    
    \item We introduce \textsc{\benchmark}, a dedicated benchmark for evaluating reward models in tool-use settings. Our analysis reveals that even state-of-the-art LLMs and specialized reward models show significant gaps on this benchmark, underscoring the need for targeted solutions.
\end{itemize}

\section{Related Work}
\label{sec:related_work}
\paragraph{Tool Learning in the Era of LLMs.}
Early work on agentic AI, such as \citet{yao2023react}, combines chain-of-thought reasoning~\citep{wei2022chain} with tool-augmented actions to elicit LLMs' tool-use capabilities. Later methods imitate curated tool-use trajectories via supervised fine-tuning~\citep{schick2023toolformer,liu2024apigen}, but often struggle with out-of-distribution tasks. More recent works integrate verified rewards into tool-aware reasoning, with designs tailored for question answering~\citep{jin2025searchr,song2025r1}, math~\citep{feng2025retool,dong2025tool}, and general tool-use~\citep{qian2025toolrl,zhang2025nemotron}. \citet{agarwal2025toolrm} evaluate reward models on function-calling tasks but omits multi-turn scenarios and rely on data pairs too simple for powerful models to distinguish.
\paragraph{Reward Modeling.}
Reward models guide large language models toward human-preferred outputs~\citep{ouyang2022training,bai2022constitutional}. They are typically either (1) discriminative (DiscRM), outputting scalar scores to rank responses~\citep{cai2024internlm2,liu2024skywork}, or (2) generative (GenRM), producing textual rewards for domains such as chat~\citep{skyworkcritic2024}, code~\citep{mcaleese2024llm}, and literary translation~\citep{pombal2025m}. A recent trend views reward modeling as a reasoning process~\citep{chen2025rm,guo2025reward} to enhance reward quality.
Recent work~\citep{agarwal2025toolrm} examines RM performance on function-calling tasks, but the benchmark is not sufficiently challenging for powerful models.

\section{Methodology}
\label{sec:method}
We introduce a pipeline for training {\model}. As shown in Figure~\ref{fig:framework}, we first label tool-calling trajectories with rule-based verifiers, then construct pairwise preference data via balanced multidimensional sampling. {\model}s are trained as either GenRM or DiscRM under distinct objectives. 

\subsection{Trajectory Preparation}
\paragraph{Task Sourcing.}
To build a diverse dataset, we collate function-calling tasks from seven open-source, tool-learning datasets, spanning a wide variety of task domains and trajectory patterns: APIGen~\citep{liu2024apigen}, APIGen-MT~\citep{prabhakar2025apigen}, BUTTON~\citep{chen2025facilitating}, ComplexFuncBench~\citep{zhong2025complexfuncbench}, Hermes-Function-Calling~\citep{teknium2025hermes}, Glaive-Function-Calling\footnote{We use a 5k cleaned glaive-function-calling subset in hermes-function-calling-v1.}, and ToolAlpaca~\citep{tang2023toolalpaca}. To address format inconsistencies across these sources, we standardize all conversation records of raw tasks into format-aligned trajectories $\mathcal{T}_\text{raw} = \{\tau_{i}\}_{i=1}^N$, discarding any data with invalid role orders. The message format within each trajectory $\tau_i$ is normalized to adhere to the Hermes Function Calling standard, where special tags \texttt{<tools>}, \texttt{<tool\_call>}, and \texttt{<tool\_response>} are used to enclose tool schemas, calls, and responses, respectively. At the beginning of each $\tau_i$, a function-calling prompt is uniformly included as the system message, along with the schemas of available tools in the task. Additional agent policies are prepended to this message for complex tasks from specific sources (e.g., APIGen-MT). See Appendix~\ref{app:example_trajectory} for an example of a tool-use task trajectory.

\paragraph{Trajectory Segmentation and Validation.}
To enable subsequent rule-based verification of arbitrary trajectories against ground-truth answers, we first perform tool schema validation for each trajectory $\tau_i$. Tool schemas are typically provided as dictionary objects, which we verify as valid JSON schemas describing tools compatible with OpenAI's tool-calling format. Invalid schemas are corrected, and duplicates are removed. The validated schemas are then wrapped into function-type JSON objects and incorporated into the aforementioned system message as tool descriptions.

Next, we partition each raw trajectory $\tau_{i} \in \mathcal{T}_\text{raw}$ into sub-trajectories ending in an assistant message, yielding segmented trajectories $\mathcal{T}_\text{seg} = \{\tau_{j}\}_{j=1}^M$. Each segment $\tau_j$ consists of a conversation history $\mathbf{x}_j$ (all messages preceding the assistant message) and the corresponding assistant response $y_j$. A preliminary filtering is then applied: we retain $\tau_j$ only if the message following $y_j$ in the raw trajectory $\tau_i$ does not contain any unsuccessful tool response, ensuring the basic validity of tool calls in $y_j$.

A stricter tool-call validation is then applied to $y_j$ for each retained trajectory $\tau_{j}$. Each tool call must be parsable in the required format (e.g., \texttt{\{"name":"\dots","arguments":\{\dots\}\}}), and its function name and arguments must match the tool schema. Responses containing duplicate tool calls are also discarded. Only the trajectories $\tau_j = (\mathbf{x}_j, y_j)$ that pass all format and content checks are kept. For these validated trajectories, we treat the assistant response as the ground-truth response $y^*_j$, and the clean dataset is $\mathcal{T}_\text{clean}=\{(\mathbf{x}_j,y^*_j)\}^{M'}_{j=1}$. Data statistics are reported in Appendix~\ref{app:data_statistics}.

\paragraph{Response Sampling and Verification.}
In this phase, we begin by sampling multiple model responses for each conversation history. To ensure diversity in the outputs, we select five models from three different families with varying tool-calling capabilities: Claude-3.7-Sonnet, Gemini-2.5-Pro, Qwen2.5-Max, Qwen-32B, and Qwen3-8B. For each pair $(\mathbf{x}_j, y^*_j)$ in the cleaned dataset $\mathcal{T}_\text{clean}$, the context $\mathbf{x}_j$ is sent to all five models, yielding a set of new assistant responses $\{\hat{y}_{j,k}\}^5_{k=1}$. Each sampled response $\hat{y}_{j,k}$ is then scored using a rule-based function that compares it against its corresponding ground-truth response $y^*_j$, yielding a score between 0 and 1. Unlike prior rule-based TIR approaches~\citep{qian2025toolrl}, our method for training the reward model prioritizes the correctness of tool call content (reasoning ability) over strict format adherence (instruction-following ability), since downstream applications often use varying tool call structures. Consequently, we only score $\hat{y}$ that can be successfully parsed into the expected tool-call format and discard all others.

For a given ground-truth response $y^*$ and a sampled response $\hat{y}$ (we drop indices $j,k$ for simplicity), let $\mathcal{C}^*=\{c^*_i\}^{N_\text{G}}_{i=1}$ and $\hat{\mathcal{C}}=\{\hat{c}_l\}^{N_\text{P}}_{l=1}$ denote the lists of tool calls parsed from them, respectively. Each tool call is a JSON object containing a string-typed \texttt{name} and a dictionary of \texttt{arguments}. Scoring starts with two disqualifiers: if either applies, the final score $\hat{s}$ is set to 0:
\begin{itemize}[leftmargin=*]
    \item \textit{Mismatched Number of Tool Calls.} The number of predicted tool calls does not match the number of ground-truth tool calls:
    \begin{equation}
    \label{eq:mismatch_tc_num}
    |\hat{\mathcal{C}}| \neq |\mathcal{C}^*| \Rightarrow \hat{s}=0
    \end{equation}
    
    \item \textit{Duplicated Tool Calls.} The set of predicted tool calls contains identical duplicates (both name and arguments are the same). For $\hat{c}_l, \hat{c}_m \in \hat{\mathcal{C}}$:
    \begin{equation}
    \label{eq:duplicate_tc}
    \resizebox{0.88\linewidth}{!}{$
    \exists{l \neq m} \; \text{s.t.} \, \mathtt{is\_identical}(\hat{c}_l, \hat{c}_m) \Rightarrow \hat{s}=0
    $}
    \end{equation}
\end{itemize}

If a sampled response $\hat{y}$ passes the above initial checks, a match score $s_i$ is calculated for each ground-truth tool call $c^*_i\in \mathcal{C}^*$. This score is determined by matching $c^*_i$ with the predicted tool call of the same name that achieves the highest argument similarity. Specifically:
\begin{equation}
\label{eq:match_score}
\begin{aligned}
s_i &= \max_{\hat{c} \in \hat{\mathcal{C}}}
    \mathds{1}\bigl[c^*_i\texttt{.name}=\hat{c}\texttt{.name}\bigr] \\
    &\quad\cdot\mathtt{sim}\bigl(c^*_i\texttt{.arguments},
    \hat{c}\texttt{.arguments}\bigr)
\end{aligned}
\end{equation}

where $\mathds{1}[\cdot]$ is an indicator function equal to 1 if the tool names match and 0 otherwise, so arguments are only compared when tool names align. The argument similarity function $\mathtt{sim}(\cdot)$ measures the ratio of identical key-value pairs to the total number of unique keys across both dictionaries. A key-value pair is considered identical only if the key appears in both dictionaries and the corresponding values match, with string comparisons performed in a case-insensitive manner. If both dictionaries are empty, the similarity is defined as 1.
The final rule-based score $\hat{s}$ can then be calculated as the mean of all individual match scores $s_i$, with $\hat{s}=1$ when both $y^*$ and $\hat{y}$ contain no tool calls:
\begin{equation}
\label{eq:final_score}
\hat{s}=\frac{1}{N_\text{G}}\sum^{N_\text{G}}_{i=1}{s_i}
\end{equation}

\paragraph{Difficulty-Aware Down-Sampling.}
After collecting all rule-based scores for sampled responses, we perform difficulty-aware down-sampling. This is done by grouping all sampled responses by their original context $\mathbf{x}_j$. Empirically, tasks that are either too easy or too difficult are not ideal for model training: (1) contexts for which all sampled responses have a rule-based score of 1 are discarded, as they offer no meaningful variation for model critique; (2) contexts for which no sampled response receives a rule-based score of 1 are also removed, as such cases likely contain noise in either $\mathbf{x}_j$ or $y^*_j$. We retain the remaining candidates as:
\begin{equation}
\resizebox{0.88\linewidth}{!}{$
\mathcal{D}_\text{cand} = \{(\mathbf{x}_j, y^*_j, \hat{y}_{j,k}, \hat{s}_{j,k})\, |\,\text{context $j$ passes}\}
$}
\end{equation}
Each contains the conversation history, the ground-truth response, a sampled response, and the corresponding rule-based score. This pool serves as the source for constructing preference datasets.

\subsection{Preference Data Construction}
\paragraph{Pairwise Data Construction.}
This section describes how we construct data to train the RM as a critic. Such models typically evaluate data in either a \textit{pointwise} or \textit{pairwise} manner. In preliminary experiments, pointwise generative RMs supervised by rule-based scores exhibited superficial overfitting: it mimicked the training score distribution instead of developing genuine analytical ability, a form of reward hacking that hurt performance on out-of-distribution (OOD) tasks. To mitigate this, we train RMs on pairwise critique tasks, which rely on comparative judgments rather than direct scoring.
The pairwise reward model is designed to distinguish a preferred response from a rejected one for a given context. To construct the training data for this, we sample pairs of responses from the preprocessed data pool $\mathcal{D}_\text{cand}$, where ground-truth preferences are determined by their rule-based scores. Each pair consists of a chosen response $y^{+}$ and a rejected response $y^{-}$ that shares the same context but differs in score. We traverse $\mathcal{D}_\text{cand}$ and arrange the permutations according to the above rules to get a candidate pairwise data pool:
\begin{multline}
    \mathcal{D}_\text{pair-cand} = \{(\mathbf{x}, y^*, y^{+}, y^{-}, s^{+}, s^{-})\,|\, s^{+}>s^{-}, \\
    (\mathbf{x}, y^*, y^{+}, s^{+}), (\mathbf{x}, y^*, y^{-}, s^{-}) \in \mathcal{D}_\text{cand} \}
\end{multline}

\paragraph{Balanced Multi-Dimensional Sampling.}
To enable efficient training with fewer data, we then adopt a balanced, multi-dimensional sampling strategy to select samples from $\mathcal{D}_\text{pair-cand}$ where we focus on the following three dimensions of data:
\begin{itemize}[leftmargin=*]
    \item \textit{Diversity of Data Sources.} Incorporating a diverse range of tool schemas and user queries enhances the generalizability of trained models. To this end, we aim to sample contexts from different sources in a balanced manner. For each context $\mathbf{x}$ in data, we denote its source as $\mathbf{x}\texttt{.source}$.
    
    \item \textit{Coverage of Preference Intensity.} For each pair of chosen and rejected responses, the difference in their rule-based scores reflects the intensity of the preference signal: a large difference signifies a strong preference, while a small difference suggests a weak one. To train robust reward models, our data sampling process is designed to cover this full spectrum of preference signals, from weak to strong. For each pairwise data point, we measure its preference intensity by:
    \begin{equation}
    I_\text{preference}=s^{+}-s^{-}
    \end{equation}
    
    \item \textit{Complexity of Tasks.} Challenging the reward model with more complex tasks is essential for enhancing its analytical capabilities. We calculate the complexity score of one candidate data point according to its ground-truth response $y^*$:
    \begin{equation}
    S_\text{complex} = |\mathcal{C}^*| + \sum^{N_\text{G}}_{i=1} |c^*_i\texttt{.arguments}|
    \end{equation}
    where $\mathcal{C}^*$ is the set of tool calls parsed from $y^*$. Both the number of tool calls and arguments are accumulated to measure the task complexity. Over-complicated samples ($S_\text{complex} > 50$) are filtered out for a higher success rate of rollout trajectory in the model training stage.
\end{itemize}

Guided by the above principles, we use a heuristic algorithm to select samples from $\mathcal{D}_\text{pair-cand}$ that are more efficient for model training. Specifically, we prioritize samples with higher complexity scores $S_\text{complex}$ while ensuring that the data source $\mathbf{x}\texttt{.source}$ and preference intensity $I_\text{preference}$ are as balanced as possible, resulting in a subset of pairwise data $\mathcal{D}_\text{pair-sampled} \subseteq \mathcal{D}_\text{pair-cand}$ for subsequent model training. Details of the heuristic algorithm are provided in Appendix~\ref{app:algo_BMDS}.

\subsection{Model Training}
\paragraph{Critique Task Design.}
We train generative {\model} by prompting models with critique tasks. Given a conversation history and two candidate responses, the model must thoroughly evaluate each one and select the better response, outputting its name within \texttt{<choice>} tags. We adapt instructions to each model's native style: reasoning models use a \textit{think-mode} template, embedding evaluations in their reasoning process, while non-reasoning models use a \textit{no-think-mode} template, placing evaluations explicitly in \texttt{<evaluation>} tags.
We define unified evaluation criteria to ensure consistent, comprehensive critiques and to specify which tool-invocation errors should be penalized. For each sampled data $(\mathbf{x}, y^*, y^{+}, y^{-}, s^{+}, s^{-}) \in \mathcal{D}_\text{pair-sampled}$, we format the conversation history $\mathbf{x}$ into a single string.
This string is then concatenated with the two assistant responses $y^{+}$ and $y^{-}$ to form the final input query $q$. To reduce position bias and prevent reward hacking during training, we randomly swap the order of the assistant responses in 50\% of the queries, recording the position of $y^{+}$ as the ground-truth answer $a$. The resulting dataset $\mathcal{D}_\text{pref} = \{(q, y^+, y^-, a)_{i}\}_{i=1}^{K}$ is then used to train the reward model. See Appendix~\ref{app:prompt_templates} for detailed prompt templates.

\paragraph{Training Objectives.}
We train generative {\model} ({\model}-Gen) in a RLVR paradigm using Group Relative Policy Optimization~\cite[GRPO,][]{shao2024deepseekmath}, a variant of Proximal Policy Optimization~\cite[PPO,][]{schulman2017proximal} that improves efficiency and reduces computational cost by replacing the critic network with grouped relative advantages. Given an input query $q$ and its ground-truth answer $a$, let $\mathcal{O}=\{o_{1}, o_{2}, \dots, o_{G}\}$ denote the set of rollout trajectories generated by the old policy $\pi_{\theta_{\text{old}}}$. Our goal is to optimize the policy $\pi_{\theta}$ by maximizing the following objective:
\begin{equation}
\resizebox{0.92\linewidth}{!}{
  $\begin{aligned}
   J_{\text{GRPO}}(\theta) ={}& \mathbb{E}_{(q,a) \sim \mathcal{D}_\text{pref}, \{o_i\}_{i=1}^G \sim \pi_{\theta_{\text{old}}}(\cdot|q)} \\
   & \left[ \frac{1}{G} \sum_{i=1}^G \frac{1}{|o_i|} \sum_{t=1}^{|o_i|} \left[ \min \left( \frac{\pi_{\theta}(o_{i,t}|q, o_{i,<t})}{\pi_{\theta_{\text{old}}}(o_{i,t}|q, o_{i,<t})} A_{i,t}, \right. \right. \right. \\
   & \left. \left. \left. \operatorname{clip} \left( \frac{\pi_{\theta}(o_{i,t}|q, o_{i,<t})}{\pi_{\theta_{\text{old}}}(o_{i,t}|q, o_{i,<t})}, 1-\epsilon, 1+\epsilon \right) A_{i,t} \right) \right] \right]
  \end{aligned}$
}
\end{equation}
where $\epsilon$ is a clipping-related hyper-parameter for stabilizing training. $A_{i,t}$ denotes the relative advantage calculated on outputs of each rollout group:
\begin{equation}
    A_{i,t} = \frac{r_{i}-\operatorname{mean}(\{r_1, r_2, \dots, r_G\})}{\operatorname{std}(\{r_1, r_2, \dots, r_G\})}
\end{equation}
Here, $r_{i}$ denotes the binary reward assigned to the rollout trajectory $o_{i}$. It is determined by whether a valid choice can be successfully extracted from $o_{i}$ and whether it accurately answers $q$:

{\small
\[
r_{i} = 
\begin{cases}
    1, \text{if} \ \mathtt{is\_equivalent}(a, \mathtt{extract\_choice}(o_{i})))\\
    0, \text{otherwise}.
\end{cases}
\]
}
Following~\citet{qian2025toolrl}, we omit the KL penalty term from the original GRPO objective to encourage more effective exploitation of reward signals during policy updates. Building on this, we design a verifiable reward system for training generative reward models in the tool-use scenario.

We train discriminative {\model} ({\model}-Disc) with a pairwise ranking loss based on the Bradley-Terry model~\citep{bradley1952rank}, following \citet{ouyang2022training}:
\begin{equation}
    \mathcal{L}_{\text{ranking}} = -\log (\sigma (r_{\theta} (\mathbf{x}, y^{+}) - r_{\theta} (\mathbf{x}, y^{-})))
\end{equation}
where $r_\theta(\mathbf{x}, y^{+})$ and $r_\theta(\mathbf{x}, y^{-})$ are the scalar rewards assigned by the reward model $\theta$ to the chosen response $y^{+}$ and the rejected response $y^{-}$ for a given prompt $\mathbf{x}$.

\section{Experiments}
\subsection{Do ToolRM Provide Precise Rewards?}
\paragraph{Benchmark Construction.}
We evaluate reward models on an improved benchmark adapted from \citet{agarwal2025toolrm}, based on BFCL. The original benchmark pairs correct function calls with incorrect ones generated by 25 permissively licensed models but has two main limitations: (1) it only covers single-turn tasks, and (2) its data pairs are too trivial for powerful RMs to differentiate.
To address this, we build a more challenging benchmark using the \textit{multi\_turn\_base} split from BFCL-v3 and curate harder rejected responses from seven top-performing function-calling models.
\begin{table*}[!ht]
\centering
\renewcommand{\arraystretch}{1.3}
\caption{Evaluation results of reward models on {\benchmark}. Higher accuracy indicates a stronger ability to distinguish better responses. GenRMs and DiscRMs trained on \textit{\dataset} are highlighted in \colorbox{lightgreen}{green}. The best result in each group is \textbf{bolded}, and the second-best is \underline{underlined}. ({\scriptsize $\diamondsuit$}): evaluated with the \textit{think-mode} template; ({\scriptsize $\heartsuit$}): evaluated with the \textit{no-think-mode} template; ({\scriptsize $\clubsuit$}): evaluated with the official template. ({\scriptsize \coloremojicode{2696}}): pairwise inputs; ({\scriptsize \coloremojicode{1F3AF}}): pointwise inputs; ({\scriptsize \coloremojicode{1F4AC}}): critique as output; ({\scriptsize \coloremojicode{2705}}): choice as output; ({\scriptsize \coloremojicode{1F522}}): scalar reward as output.}
\scalebox{0.65}{
\renewcommand{\arraystretch}{0.78}
\begin{tabular}{lccccccccccc}
\toprule
& \multicolumn{11}{c}{Classification Accuracy (\%)} \\
\cline{2-12}
\multirow{-2}{*}{Models} & S & M & P & PM & LS & LM & LP & LPM & MTB & \textbf{Avg.} & \textbf{W-Avg.} \\
\midrule

\multicolumn{12}{c}{\textit{Proprietary \& Open-source General LLMs}} \\
\midrule
\coloremojicode{2696}\coloremojicode{1F4AC} \href{https://huggingface.co/meta-llama/Llama-3.2-3B-Instruct}{Meta/Llama-3.2-3B-Instruct}$^{\heartsuit}$ & 34.31 & 33.80 & 24.58 & 34.87 & 26.89 & 29.54 & 8.82 & 30.00 & 20.20 & 27.00 & 28.09 \\
\coloremojicode{2696}\coloremojicode{1F4AC} \href{https://huggingface.co/meta-llama/Llama-3.1-8B-Instruct}{Meta/Llama-3.1-8B-Instruct}$^{\heartsuit}$ & 45.99 & 52.11 & 46.84 & 62.31 & 33.02 & 39.44 & 23.53 & 40.00 & 28.69 & 41.33 & 41.38 \\
\coloremojicode{2696}\coloremojicode{1F4AC} \href{https://huggingface.co/Qwen/Qwen3-4B-Thinking-2507}{Qwen/Qwen3-4B-Thinking-2507}$^{\diamondsuit}$ & 67.88 & 70.42 & 85.71 & 87.69 & 61.79 & 46.61 & \textbf{85.29} & \underline{85.00} & 33.54 & 69.33 & 57.59 \\
\coloremojicode{2696}\coloremojicode{1F4AC} \href{https://huggingface.co/Qwen/Qwen3-4B}{Qwen/Qwen3-4B (w/ thinking)} $^{\diamondsuit}$ & 70.07 & 73.24 & 89.70 & 87.69 & 56.60 & 48.09 & 79.41 & 81.67 & 39.80 & 69.59 & 59.34 \\
\coloremojicode{2696}\coloremojicode{1F4AC} \href{https://huggingface.co/Qwen/Qwen3-8B}{Qwen/Qwen3-8B (w/ thinking)} $^{\diamondsuit}$ & 71.53 & 61.97 & 89.37 & 90.26 & 58.49 & 48.09 & \textbf{85.29} & 76.67 & 39.19 & 68.98 & 59.44 \\
\coloremojicode{2696}\coloremojicode{1F4AC} \href{https://huggingface.co/Qwen/Qwen3-4B-Instruct-2507}{Qwen/Qwen3-4B-Instruct-2507}$^{\heartsuit}$ & 71.53 & 64.79 & 90.37 & 89.23 & 51.42 & 50.66 & 70.59 & \textbf{86.67} & 36.57 & 67.98 & 59.67 \\
\hdashline
\coloremojicode{2696}\coloremojicode{1F4AC} DeepSeek-AI/DeepSeek-R1-0528$^{\diamondsuit}$ & 68.61 & 70.42 & 87.71 & 85.64 & \underline{64.62} & 46.45 & 76.47 & 75.00 & 36.77 & 67.97 & 57.93 \\
\coloremojicode{2696}\coloremojicode{1F4AC} OpenAI/GPT-4o-2024-11-20$^{\heartsuit}$ & 69.34 & 66.20 & 86.71 & 86.67 & 50.47 & 50.82 & 67.65 & 78.33 & 38.38 & 66.06 & 59.00 \\
\coloremojicode{2696}\coloremojicode{1F4AC} OpenAI/o3-2025-04-16$^{\diamondsuit}$ & 70.80 & 69.01 & 85.71 & 84.87 & 55.19 & 50.43 & 67.65 & 76.67 & \textbf{41.21} & 66.84 & 59.40 \\
\coloremojicode{2696}\coloremojicode{1F4AC} Google/Gemini-2.5-Flash (w/ thinking)$^{\diamondsuit}$ & 64.23 & 66.20 & 89.70 & 89.49 & 56.13 & 51.13 & 79.41 & 80.00 & 36.77 & 68.12 & 59.87 \\
\coloremojicode{2696}\coloremojicode{1F4AC} Google/Gemini-2.5-Pro (w/ thinking)$^{\diamondsuit}$ & 75.18 & 67.61 & 88.04 & \underline{91.79} & 58.96 & 48.32 & \underline{82.35} & 73.33 & 39.80 & 69.49 & 59.94 \\
\coloremojicode{2696}\coloremojicode{1F4AC} Qwen/Qwen3-235B-A22B-Thinking-2507$^{\diamondsuit}$ & 71.53 & 69.01 & 86.05 & 90.26 & \textbf{67.92} & 51.52 & \textbf{85.29} & 76.67 & 34.55 & 70.31 & 60.64 \\
\coloremojicode{2696}\coloremojicode{1F4AC} DeepSeek-AI/DeepSeek-V3-0324$^{\heartsuit}$ & 75.18 & 66.20 & 88.70 & 89.74 & 58.02 & 53.86 & 70.59 & 73.33 & 37.17 & 68.09 & 61.45 \\
\coloremojicode{2696}\coloremojicode{1F4AC} Qwen/Qwen2.5-Max$^{\heartsuit}$ & \underline{77.37} & \underline{73.24} & 89.04 & 90.00 & 58.02 & \textbf{55.18} & 67.65 & 70.00 & 37.98 & 68.72 & 62.39 \\
\coloremojicode{2696}\coloremojicode{1F4AC} Anthropic/Claude-3.7-Sonnet (w/ thinking)$^{\diamondsuit}$ & 76.64 & 67.61 & \textbf{91.69} & \textbf{92.82} & 60.85 & 52.77 & 73.53 & 78.33 & 39.19 & \underline{70.38} & \underline{62.45} \\
\coloremojicode{2696}\coloremojicode{1F4AC} \textbf{Anthropic/Claude-4-Sonnet (w/ thinking)}$^{\diamondsuit}$ & \textbf{81.02} & \textbf{77.46} & \underline{91.36} & 91.28 & 62.74 & \underline{54.95} & \underline{82.35} & 83.33 & \underline{41.01} & \textbf{73.95} & \textbf{64.23} \\
\midrule
\multicolumn{12}{c}{\textit{Generative Reward Models}} \\
\midrule
\coloremojicode{1F3AF}\coloremojicode{1F4AC}\coloremojicode{1F522} \href{https://huggingface.co/ankner/Llama3-8B-CLoud-RM}{Databricks/CLoud-RM-Llama-3-8B}$^{\clubsuit}$ & 25.55 & 35.21 & 33.22 & 32.82 & 31.60 & 37.88 & 32.35 & 25.00 & 49.90 & 33.73 & 37.34 \\
\coloremojicode{2696}\coloremojicode{1F4AC} \href{https://huggingface.co/Unbabel/M-Prometheus-7B}{Unbabel/M-Prometheus-7B}$^{\heartsuit}$ & 54.74 & 54.93 & 71.43 & 74.87 & 43.87 & 46.69 & 38.24 & 53.33 & 34.14 & 52.47 & 51.19 \\
\coloremojicode{2696}\coloremojicode{1F4AC} \href{https://huggingface.co/Reward-Reasoning/RRM-7B}{Microsoft-Research/RRM-7B}$^{\diamondsuit}$ & 65.69 & 56.34 & 82.06 & 84.62 & 43.40 & 49.65 & 44.12 & 68.33 & 36.36 & 58.95 & 56.05 \\
\coloremojicode{2696}\coloremojicode{1F4AC} \href{https://huggingface.co/gaotang/RM-R1-DeepSeek-Distilled-Qwen-32B}{UIUC/RM-R1-DeepSeek-Distilled-Qwen-32B}$^{\diamondsuit}$ & 75.18 & 76.06 & 68.44 & 80.51 & 61.79 & 49.18 & 52.94 & 53.33 & 38.18 & 61.73 & 56.25 \\
\coloremojicode{2696}\coloremojicode{1F4AC} \href{https://huggingface.co/Unbabel/M-Prometheus-14B}{Unbabel/M-Prometheus-14B}$^{\heartsuit}$ & 64.96 & 57.75 & 88.37 & 87.44 & 44.34 & 46.38 & 64.71 & 61.67 & 39.39 & 61.67 & 56.32 \\
\coloremojicode{2696}\coloremojicode{2705} \href{https://huggingface.co/Skywork/Skywork-Critic-Llama-3.1-8B}{Skywork/Skywork-Critic-Llama-3.1-8B}$^{\clubsuit}$ & 54.74 & 59.15 & 86.05 & 83.59 & 47.17 & 45.75 & 67.65 & 61.67 & 50.30 & 61.79 & 56.92 \\
\rowcolor{lightgreen}
\coloremojicode{2696}\coloremojicode{1F4AC} \textsc{\model}-Gen-Llama-3.2-3B-Instruct$^{\heartsuit}$ & 54.01 & 57.75 & 87.04 & 78.97 & 44.34 & 54.95 & 64.71 & 61.67 & 45.45 & 60.99 & 59.27 {\small \textcolor{darkgreen}{(+31.18)}} \\
\rowcolor{lightgreen}
\coloremojicode{2696}\coloremojicode{1F4AC} \textsc{\model}-Gen-Llama-3.1-8B-Instruct$^{\heartsuit}$ & 62.04  & 61.97 & 88.70 & 86.15 & 47.64 & 52.30 & 82.35 & 68.33 & 41.21 & 65.63 & 59.57 {\small \textcolor{darkgreen}{(+18.19)}} \\
\coloremojicode{2696}\coloremojicode{2705} \href{https://huggingface.co/Skywork/Skywork-Critic-Llama-3.1-70B}{Skywork/Skywork-Critic-Llama-3.1-70B}$^{\clubsuit}$ & 64.23 & 67.61 & 87.38 & 88.21 & 44.34 & 51.68 & 70.59 & 66.67 & 47.47 & 65.35 & 60.31 \\
\coloremojicode{2696}\coloremojicode{1F4AC} \href{https://huggingface.co/Reward-Reasoning/RRM-32B}{Microsoft-Research/RRM-32B}$^{\diamondsuit}$ & 76.64 & 76.06 & 87.38 & 89.23 & 67.92 & 56.90 & 67.65 & 75.00 & 42.83 & 71.07 & 64.50 \\
\rowcolor{lightgreen}
\coloremojicode{2696}\coloremojicode{1F4AC} \textsc{\model}-Gen-Qwen3-4B-Instruct-2507$^{\heartsuit}$ & 70.80 & 74.65 & \textbf{91.03} & 89.49 & 55.66 & 60.41 & \textbf{94.12} & 81.67 & 49.90 & 74.19 & 66.85 {\small \textcolor{darkgreen}{(+7.18)}} \\
\rowcolor{lightgreen}
\coloremojicode{2696}\coloremojicode{1F4AC} \textsc{\model}-Gen-Qwen3-4B$^{\diamondsuit}$ & \underline{81.02} & \underline{78.87} & 89.04 & 88.97 & 63.21 & \underline{62.12} & \underline{91.18} & \underline{86.67} & \underline{52.32} & \underline{77.04} & 68.89 {\small \textcolor{darkgreen}{(+9.55)}} \\
\rowcolor{lightgreen}
\coloremojicode{2696}\coloremojicode{1F4AC} \textsc{\model}-Gen-Qwen3-8B$^{\diamondsuit}$ & \underline{81.02} & 76.06 & 89.70 & \underline{91.03} & \underline{64.62} & 61.50 & \underline{91.18} & 80.00 & \textbf{52.73} & 76.43 & \underline{68.92} {\small \textcolor{darkgreen}{(+9.48)}} \\
\rowcolor{lightgreen}
\coloremojicode{2696}\coloremojicode{1F4AC} \textbf{\textsc{\model}-Gen-Qwen3-4B-Thinking-2507}$^{\diamondsuit}$ & \textbf{83.21} & \textbf{80.28} & \underline{90.03} & \textbf{92.56} & \textbf{71.23} & \textbf{66.02} &  \textbf{94.12} & \textbf{88.33} & 52.12 & \textbf{79.77} & \textbf{71.87 {\small \textcolor{darkgreen}{(+14.28)}}} \\

\midrule
\multicolumn{12}{c}{\textit{Discriminative Reward Models}} \\
\midrule
\coloremojicode{1F3AF}\coloremojicode{1F522} \href{https://huggingface.co/Skywork/Skywork-Reward-Llama-3.1-8B-v0.2}{Skywork/Skywork-Reward-Llama-3.1-8B-v0.2}$^{\clubsuit}$ & 83.21 & 70.42 & 92.36 & \textbf{92.31} & 59.91 & 62.51 & 67.65 & 75.00 & 59.80 & 73.68 & 70.23 \\
\coloremojicode{1F3AF}\coloremojicode{1F522} \href{https://huggingface.co/internlm/internlm2-7b-reward}{InternLM/InternLM2-7B-Reward}$^{\clubsuit}$ & 80.29 & \underline{80.28} & 88.04 & 89.74 & 63.68 & 65.16 & 67.65 & 73.33 & \underline{61.21} & 74.38 & 71.17 \\
\coloremojicode{1F3AF}\coloremojicode{1F522} \href{https://huggingface.co/Skywork/Skywork-Reward-V2-Llama-3.1-8B}{Skywork/Skywork-Reward-V2-Llama-3.1-8B}$^{\clubsuit}$ & \textbf{88.32} & 77.46 & 90.70 & 91.03 & 68.40 & 64.54 & 70.59 & \textbf{90.00} & 60.81 & 77.98 & 72.28 \\
\coloremojicode{1F3AF}\coloremojicode{1F522} \href{https://huggingface.co/internlm/internlm2-20b-reward}{InternLM/InternLM2-20B-Reward}$^{\clubsuit}$ & \underline{87.59} & \textbf{84.51} & 91.69 & 91.54 & 64.15 & 68.67 & 88.24 & 76.67 & 55.35 & 78.71 & 73.08 \\
\coloremojicode{1F3AF}\coloremojicode{1F522} \href{https://huggingface.co/Skywork/Skywork-Reward-V2-Qwen3-4B}{Skywork/Skywork-Reward-V2-Qwen3-4B}$^{\clubsuit}$ & 85.40 & \underline{80.28} & \textbf{94.35} & \underline{92.05} & \underline{70.28} & 67.58 & 76.47 & \underline{83.33} & 55.76 & 78.39 & 73.25 \\
\rowcolor{lightgreen}
\coloremojicode{1F3AF}\coloremojicode{1F522} \textsc{\model}-Disc-Qwen3-4B-Thinking-2507 & 81.75 & \textbf{84.51} & 93.69 & 90.26 & \textbf{72.17} & \textbf{74.51} & \underline{97.06} & 71.67 & 60.61 & \underline{80.69} & \underline{76.80} {\small \textcolor{darkgreen}{(+19.21)}} \\
\rowcolor{lightgreen}
\coloremojicode{1F3AF}\coloremojicode{1F522} \textbf{\textsc{\model}-Disc-Qwen3-4B-Instruct-2507} & 84.67 & \textbf{84.51} & \underline{94.02} & 90.00 & \textbf{72.17} & \underline{73.97} & \textbf{100.00} & 80.00 & \textbf{64.85} & \textbf{82.69} & \textbf{77.61 {\small \textcolor{darkgreen}{(+17.94)}}} \\


\bottomrule
\label{tab:main_results}
\end{tabular}
}

\end{table*}

The resulting benchmark, \textsc{\benchmark}, comprises 2,983 samples from 1,397 unique tasks across 9 splits: simple (S), multiple (M), parallel (P), parallel multiple (PM), live sample (LS), live multiple (LM), live parallel (LP), live parallel multiple (LPM), and multi-turn base (MTB). It covers 20 distinct error types with rejected responses from 7 different models. Since BFCL tasks and their synthetic data are excluded from training, \textbf{\textsc{\benchmark} serves as a strong OOD evaluation set for \textsc{\model}}. Additional statistics and implementation details are in Appendix~\ref{app:benchmark_implementation}.

\paragraph{Evaluation Metric.}
We assess model performance via pairwise preference classification. To minimize position bias, each sample is evaluated twice, swapping the response order on the second pass. A sample is correct only if both orders yield the correct prediction. For scalar-output RMs, we compute scores for chosen and rejected responses and mark the result correct if the score order matches the preference label. We report average accuracy (\textbf{Avg.}) across splits and weighted-average accuracy (\textbf{W-Avg.}), based on sample counts.

\paragraph{Model Training.}
\label{sec:exp_model_training}
We train {\model} on three reasoning models (Qwen3-4B, Qwen3-8B, and Qwen3-4B-Thinking-2507) and three non-reasoning models (Qwen3-4B-Instruct-2507, Llama-3.2-3B-Instruct, and Llama-3.1-8B-Instruct~\citep{dubey2024llama}). At both training and inference for generative {\model}, we apply the appropriate \textit{think-mode} or \textit{no-think-mode} templates. Our preference dataset, \textit{\dataset}, contains 30,000 samples (29,500 for training, 500 for validation), built with our proposed pipeline. See training details in Appendix~\ref{app:model_training} and impact of data scaling on {\model} in Appendix~\ref{app:data_scaling}.

\paragraph{Baseline Models.}
We benchmark {\model} against strong LLMs in the LLM-as-a-judge setup, including GPT, Gemini, Claude, DeepSeek, and Qwen. Specialized reward models are also tested: generative (Skywork-Critic~\citep{skyworkcritic2024}, M-Prometheus~\citep{pombal2025m}, RM-R1~\citep{chen2025rm}, RRM~\citep{guo2025reward}), discriminative (Skywork-Reward~\citep{liu2024skywork,liu2025skywork}, InternLM2-Reward~\citep{cai2024internlm2}), and hybrid (Cloud-RM~\citep{ankner2024critique}).

\paragraph{Main Results.}
Table~\ref{tab:main_results} presents evaluation results on \textsc{\benchmark} across all splits. Training on \textit{\dataset} significantly boosts performance, with Qwen3 models gaining 12.94\% on average and up to 17.94\% in weighted accuracy. {\model}, trained on Qwen3-4B-Thinking-2507 and Qwen3-4B-Instruct-2507, achieves the best results within each model group. Notably, {\model} also improves on the \textit{multi-turn-base} split, despite being trained on step-wise critiques. Since BFCL scoring for multi-turn tasks relies on state- and response-based signals rather than rule-matching, these gains demonstrate that \textbf{{\model} acquires robust, generalizable analytical capabilities rather than overfitting to rule-based labels.}

In LLM-as-a-judge evaluations, Claude-4 outperforms other general-purpose LLMs, consistent with its stronger tool-use capabilities. Among specialized reward models, the Skywork-Reward-V2 series performs best, likely due to training on 26M diverse preference pairs. Notably, Skywork-Reward-Llama-3.1-8B-v0.2 exceeds its generative counterpart, Skywork-Critic, despite similar training data—a pattern also observed in {\model}. This suggests that discriminative reward models may generalize better to pairwise classification tasks than generative critics (see Appendix~\ref{app:ablation_model_training} for discussion).

Lastly, reasoning models show greater gains from pairwise RL training than instruction-tuned counterparts, and models with longer initial reasoning patterns (e.g., Qwen3-4B-Thinking-2507 vs. Qwen3-4B) benefit the most. This highlights that \textbf{even with weaker initial performance, a greater capacity for exploration can ultimately lead to stronger outcomes through RL.} See the ablation on training data construction in Appendix~\ref{app:ablation_data_construction}.

\begin{figure}[!t]
    \centering
    \begin{minipage}{0.44\columnwidth}
        \centering
        \includegraphics[width=\linewidth]{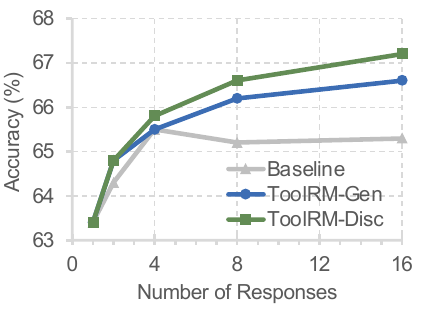}
        \caption{Model BoN sampling on ACEBench.}
        \label{fig:acebench_BoN}
    \end{minipage}
    \hfill
    \begin{minipage}{0.52\columnwidth}
        \centering
        \includegraphics[width=\linewidth]{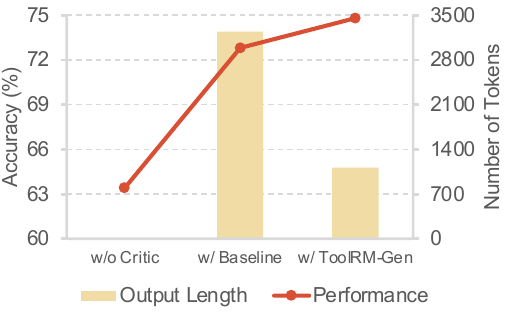}
        \caption{Model self-correction on ACEBench.}
        \label{fig:acebench_critique}
    \end{minipage}
\end{figure}

\subsection{Do {\model} Aid Inference-Time Scaling?}
\paragraph{Setup.}
We assess whether {\model} improves tool-call inference using 823 samples from the \textit{Normal} split of \textsc{ACEBench}~\citep{chen2025acebench}, a benchmark for tool-use evaluation. For each sample, we apply Best-of-N (BoN) sampling with Qwen3-4B-Instruct-2507 (temperature = 1.0), and use trained RMs to select the best response. We compare three judges: the baseline Qwen3-4B-Thinking-2507 (\textit{Base}), the trained {\model}-Gen-Qwen3-4B-Thinking-2507 (\textit{ToolRM-Gen}), and {\model}-Disc-Qwen3-4B-Instruct-2507 (\textit{ToolRM-Disc}). Performance is measured by average accuracy across all samples.

\paragraph{Main Results.}
Figure~\ref{fig:acebench_BoN} shows that {\model} (-Disc/-Gen) consistently matches or outperforms the baseline, improving by 3.8/3.2 points without BoN and 1.9/1.3 points with BoN-16. These gains validate the quality of \textit{ToolPref} and indicate that ToolRM-Gen generalizes beyond its RL training objective. Notably, performance remains stable as the candidate pool grows, demonstrating \textbf{robustness to long-context reasoning} and \textbf{effectiveness for inference-time scaling in tool-use tasks}.

\subsection{Are Model Critiques Helpful?}

\paragraph{Setup.}
We evaluate how critiques generated by {\model} guide policy model self-correction (SC). For each sample in the \textit{Normal} subset of ACEBench, Qwen3-4B-Instruct-2507 first produces a function-calling response. A GenRM then critiques the output with concise feedback, which the policy model uses to refine its response. We compare two critics here: the baseline Qwen3-4B-Thinking-2507 (\textit{Baseline}) and the trained {\model}-Gen-Qwen3-4B-Thinking-2507 (\textit{ToolRM-Gen}). Performance is measured by average accuracy over all samples.

\paragraph{Main Results.}
As shown in Figure~\ref{fig:acebench_critique}, {\model}-Gen leads to notable gains in self-correction accuracy: +11.4 points over \textit{w/o Critic} and +2.0 over \textit{w/ Baseline}, indicating more reliable critiques in tool use tasks. It also lowers decoding cost, reducing average output length from 3,211 to 1,111 tokens, demonstrating \textbf{efficient reasoning without sacrificing critique quality.} See Appendix~\ref{app:case_study} for qualitative examples and Appendix~\ref{app:ablation_model_training} for a detailed comparison of ToolRM-Gen and ToolRM-Disc.

\begin{table}[!t]
    \centering
    \caption{Policy model accuracy (\%) on BFCL-v3 before and after RL training with {\model}-Gen.}
    \label{tab:model_rl_res}
    \scalebox{0.7}{
        \begin{tabular}{lccc}
            \toprule
            \textbf{Model} & \textbf{Single-Turn AST} & \textbf{Multi-Turn} \\
            \midrule
            Qwen3-4B-Instruct-2507 & 73.25 & 19.88 \\
            \rowcolor{lightgreen}
            - \textit{RL w/ ToolRM-Gen} & \textbf{77.89 {\small \textcolor{darkgreen}{(+4.64)}}} & \textbf{25.50 {\small \textcolor{darkgreen}{(+5.62)}}} \\
            \bottomrule
        \end{tabular}
    }
\end{table}

\subsection{Do {\model} Facilitate Policy RL Training?}
\paragraph{Setup.}
We test whether {\model} can improve another policy model's tool use by serving as a reliable reward model during RL training. Using 15,000 unlabeled tool-call queries, we train Qwen3-4B-Instruction-2507 with GRPO. During training, the policy model's rollouts are paired and scored by {\model}-Gen, which assigns a reward of 1 to the chosen response and 0 to the other. We then evaluate the resulting policy model on BFCL-v3.

\paragraph{Main Results.}
Table~\ref{tab:model_rl_res} reports policy-model accuracy on BFCL-v3 across splits and shows substantial gains after supervised RL with {\model}-Gen. The fact that applying the reward model to downstream RL enhances tool-calling agents further \textbf{underscores the practical utility of {\model}}. See Appendix~\ref{app:model_rl_setup} for detailed settings and results.

\subsection{Error Analysis of {\model}}
\label{sec:error_analysis}
We further analyze the thinking process of generative {\model} in cases where its final judgments are inconsistent with the ground-truth preferences. Our investigation indicates that these errors primarily fall into two categories: (i) when the description of tool schema or parameters lacks concrete examples, the model is unable to infer the most appropriate tool invocation from the candidates, given the available tool information and the user's query; (ii) the originally annotated chosen response contains minor errors, while the rejected response has more fundamental and severe errors. The model correctly identifies all errors but fails to distinguish primary errors from secondary ones, leading to an incorrect pairwise reward.
We believe the first type of error is constrained by the base model's inherent reasoning capability and is therefore more difficult to improve. The second type, however, is more tractable and can be mitigated through targeted optimization using higher-quality, non-perfect preference pairs, of which the chosen response still contains minor errors. Examples from {\benchmark} corresponding to the two typical error types are provided in Appendix~\ref{app:case_study} for detailed reference.

\section{Conclusion}
\label{sec:conclusion}
This paper introduces a framework of data, models and benchmarks for agentic tool-use reward modeling. At its core is a novel data construction pipeline that combines rule-based labeling with balanced multi-dimensional sampling to automatically generate fine-grained pairwise preference data. The resulting dataset is diverse, well-balanced, and intentionally challenging, enabling efficient RL training and promoting nuanced reasoning beyond superficial signal matching.

Extensive evaluations across multiple benchmarks demonstrate the value of {\model} in: (i) delivering high-fidelity reward signals that align with human preferences and outperform frontier baselines; (ii) enabling inference-time scaling by reliably selecting optimal outputs from diverse candidate pools; (iii) producing efficient, pointwise critiques that improve self-correction with minimal decoding overhead; and (iv) effectively supporting downstream RL training.

Together, these results suggest that reward models trained on high-quality data with suitable objectives can serve as effective judges and critics to support downstream decision-making in general tool-use settings. Future work could extend this approach to more open-ended agentic tasks, including RM-guided multi-agent coordination and planning. We hope this work provides new perspectives on efficient tool learning for LLMs.


\section*{Limitations}
Although the trained {\model} generalizes across tasks, it still has limitations. In agentic tool-use scenarios, whether {\model} should be viewed as an outcome reward model (ORM) or process reward model (PRM) depends on the application context: (i) Single-action view: {\model} scores each action's final output without access to intermediate chain-of-thought, so it effectively serves as an ORM. (ii) Multi-turn view: In longer trajectories, {\model} primarily evaluates each step independently and is not intended to judge the overall task outcome or provide holistic trajectory-level feedback, aligning it more with a PRM. Extending {\model} to evaluate full trajectories is important but technically challenging; because it would require different objectives and substantial additional work, we leave it to future research.


\bibliography{anthology-1,anthology-2,custom}


\clearpage
\appendix
\begin{table*}[!t]
\centering
\caption{Statistics for each constituent dataset. \textit{Raw} and \textit{Filtered} are reported by the number of original tasks, while \textit{Segmented} counts the number of segmented trajectories, with \textit{Msg} indicating their average number of messages. Trajectory patterns in each dataset are characterized from turn, step, and order perspectives: `ST' and `MT' denote `single-turn' and `multi-turn'; `SS' and `MS' denote `single-step' and `multi-step'; `P' and `S' denote `parallel' and `sequential', respectively.}
\label{tab:data_statistics}
\scalebox{0.63}{
\begin{tabular}{lrrrrrcccc}
\toprule
\multirow{2}{*}{\textbf{Data Source}} &
  \multicolumn{1}{c}{\multirow{2}{*}{\textbf{\#Raw}}} &
  \multicolumn{1}{c}{\multirow{2}{*}{\textbf{\#Filtered}}} &
  \multicolumn{1}{c}{\multirow{2}{*}{\textbf{\#Segmented}}} &
  \multicolumn{1}{c}{\multirow{2}{*}{\textbf{\#Msg}}} &
  \multicolumn{1}{c}{\multirow{2}{*}{\textbf{\#Schemas}}} &
  \multicolumn{3}{c}{\textbf{Pattern of Trajectory}} &
  \multicolumn{1}{c}{\multirow{2}{*}{\textbf{Task Domain}}} \\ \cline{7-9}
 &
  \multicolumn{1}{c}{} &
  \multicolumn{1}{c}{} &
  \multicolumn{1}{c}{} &
  \multicolumn{1}{c}{} &
  \multicolumn{1}{c}{} &
  \multicolumn{1}{c}{\textbf{Turn}} &
  \multicolumn{1}{c}{\textbf{Step}} &
  \multicolumn{1}{c}{\textbf{Order}} &
  \multicolumn{1}{c}{} \\ \midrule
APIGen                  & 60,000 & 60,000 & 59,960 & 3.00  & 4,205  & ST & SS/MS & P      & Finance/Sports/Technology/Travel \dots         \\
APIGen-MT               & 5,000  & 4,874  & 20,055 & 11.75 & 26     & MT & SS/MS & P/S & Airline/Retail                               \\
BUTTON                  & 8,000  & 8,000  & 20,811 & 5.19  & 22,101 & ST & SS/MS & P/S & Daily Life                                   \\
ComplexFuncBench        & 1,000  & 1,000  & 3,259  & 5.43  & 40     & ST & MS    & S       & Hotel/Flight/Attraction/Car Rental/Taxi      \\
Glaive-Function-Calling & 5,209  & 4,344  & 6,747  & 4.82  & 1,565  & MT & SS/MS & P      & Stocks and Orders/Movie/Flight Services \dots  \\
Hermes-Function-Calling & 1,893  & 1,724  & 1,724  & 3.00  & 2,383  & ST & SS/MS & P      & Information Extraction/API Call/Software \dots \\
ToolAlpaca              & 4,098  & 2,510  & 6,194  & 4.24  & 2,040  & ST & SS/MS & P/S & News/Jobs/Finance/Entertainment \dots          \\ \bottomrule
\end{tabular}}
\end{table*}

\section{The Use of Large Language Models}
During the completion of this work, we employed Gemini 2.5 Pro~\citep{comanici2025gemini} to identify grammatical errors and refine the text in the preliminary draft stage. The data construction pipeline code was initially developed by the human authors and then verified using Qwen3-Coder~\citep{yang2025qwen3}. All suggestions from the LLMs were manually reviewed and confirmed for accuracy.

\section{Statistics of Data Sources}
\label{app:data_statistics}
In this work, we utilize seven high-quality tool-use datasets with diverse distributions. Table~\ref{tab:data_statistics} summarizes statistics for each data source, including the number of unique tool schemas and the distribution of tool-call trajectory patterns, measured by turn-, step-, and order-wise occurrences.

\section{Full Related Work}
\label{app:full_related_work}
\subsection{Tool Learning in the Era of LLMs}
The emergence of foundational capabilities in large language models (LLMs) has enabled them to identify and use appropriate tools in a human-like manner. \citet{yao2023react} unlock this ability by combining chain-of-thought reasoning~\citep{wei2022chain} with tool-augmented actions. Another line of approaches clones behaviors from completed tool-calling trajectories using supervised fine-tuning~\citep{schick2023toolformer,tang2023toolalpaca,liu2024apigen,liu2025toolace}, while these methods may face challenges generalizing to complex and out-of-distribution tasks. To address this limitation, other approaches employ reinforcement learning with human preference data to learn via trial-and-error~\citep{nakano2021webgpt}. Building on recent successes in reasoning models~\citep{lambert2025tulu3,shao2024deepseekmath}, utilizing verified rewards to facilitate tool-integrated reasoning has become a promising direction. Reward designs based on the format and correctness of the final answer have proven effective in tasks like question-answering~\citep{jin2025searchr,song2025r1}, math~\citep{feng2025retool,dong2025tool}, and general tool-calling~\citep{qian2025toolrl,zhang2025nemotron,jiang2026scribe}, leading to generalized model improvements through reinforcement learning.

\subsection{Evaluation of LLM Tool-Use}
Numerous tool-calling benchmarks have been proposed in recent years.
To enable realistic and reliable evaluation, tasks are either drawn from real-world domains~\citep{wang2024gta,patil2024gorilla,zhong2025complexfuncbench,yao2025taubench,barres2025tau} or generated via well-designed data-synthesis pipelines~\citep{qin2024toolllm,chen2025acebench}.
Among these, BFCL~\citep{patil2025bfcl} covers diverse and complex patterns of tool usage and serves as a comprehensive benchmark for evaluating LLMs' tool-use capabilities. Nevertheless, there remains a lack of a benchmark that assesses whether current models can provide accurate feedback on LLM actions in tool-use scenarios. Recent work, FC-Reward-Bench~\citep{agarwal2025toolrm}, investigates reward model performance on function-calling tasks. However, it does not evaluate multi-turn tool-use scenarios, and its data pairs are too simplistic for powerful RMs to effectively distinguish between them.

\subsection{Reward Modeling of Human Preferences}
Reinforcement learning has proven effective for aligning LLMs with human preferences, using feedback from humans~\citep{ouyang2022training} or other capable LLMs~\citep{bai2022constitutional,lee2024rlaif}.
Central to this process are reward models (RMs), which are primarily developed in two ways. The first is discriminative modeling, where RMs output a scalar score to differentiate between preferred and rejected responses~\citep{yang2024qwen2,cai2024internlm2,liu2024skywork,liu2025inference}. The second is generative modeling, where models provide textual rewards as natural language critiques for tasks like chat~\citep{skyworkcritic2024,kim2024prometheus,yu-etal-2025-self}, code~\citep{mcaleese2024llm}, and literary machine translation~\citep{pombal2025m}. 
Hybrid approaches combine critiques with scalar rewards to better capture nuanced preferences~\citep{ankner2024critique,wang2025gram}, while recent work frames reward modeling as reasoning tasks~\citep{chen2025rm,wang2025gram,guo2025reward,whitehouse2025j1}. In this paper, we extend reward modeling to the agentic tool-use scenario.

Notably, there is also a line of work on tool-augmented reward modeling \cite{li2024toolaugmented,findeis-etal-2025-external,xu2025incentivizing}, which is conceptually distinct from {\model}, with different motivations and inference procedures. In our setting, {\model} is trained to evaluate another policy model's behavior on agentic tool-use tasks, and it relies solely on internal reasoning rather than invoking external tools during evaluation. By contrast, tool-augmented RMs are primarily designed for target tasks such as general QA, writing, and coding, where the policy model can complete the task without invoking any tools, and tools are instead called at evaluation time to improve the reliability of reward estimates. Consequently, these tool-augmented RMs do not apply to the scenario studied in this paper and are not directly comparable to our approach.

\begin{figure*}[!ht]
\centering
\subfloat[Distribution of data patterns.\label{fig:benchmark_data_pattern}]{%
  \includegraphics[width=0.33\textwidth]{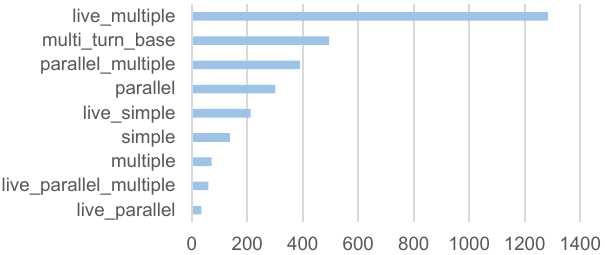}%
}\hfil
\subfloat[Distribution of error types.\label{fig:benchmark_error_type}]{%
  \includegraphics[width=0.33\textwidth]{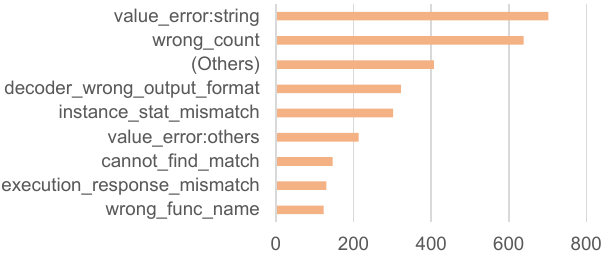}%
}
\subfloat[Distribution of response sources.\label{fig:benchmark_response_source}]{%
  \includegraphics[width=0.33\textwidth]{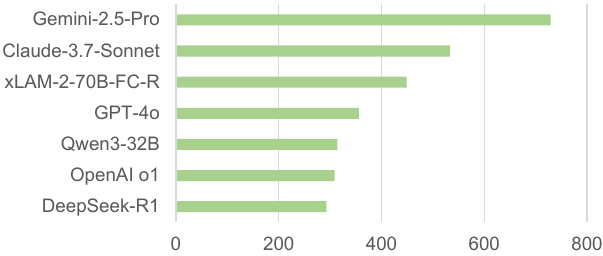}%
}
\caption{Statistics of the enhanced reward model benchmark {\benchmark}.}
\label{fig:benchmark_statistics}
\end{figure*}

\section{Experiment Details}
\label{app:experiment_details}

\subsection{Reward Model Training}
\label{app:model_training}
We train generative {\model} on eight NVIDIA A100 80G GPUs. We perform one epoch of GRPO training using veRL~\citep{sheng2025hybridflow}, with a learning rate of \texttt{1e-6} and a clip ratio of $\epsilon=0.2$. At each training step, we sample a batch of 128 queries and generate 8 trajectories per query. Trajectory generation is handled by the vllm backend~\citep{kwon2023efficient}, employing sampling hyper-parameters of \texttt{temperature=1.0}, \texttt{top\_p=1.0}, and \texttt{top\_k=-1}. Due to resource constraints, we limit the maximum prompt length to 16,384 tokens and the maximum response length to 4,096 tokens for model training.

All discriminative reward models in this paper are trained using OpenRLHF~\citep{hu-etal-2025-openrlhf} with a learning rate of \texttt{4e-6} and a training batch size of 256. For each dataset, we train for 2 epochs using the BT objective and set the maximum prompt length to 16,384 tokens during training.

\subsection{Benchmark Implementation}
\label{app:benchmark_implementation}
In constructing \textsc{\benchmark}, all rejected LLM responses are extracted from the official evaluation archive\footnote{Trajectories from \url{https://github.com/HuanzhiMao/BFCL-Result}}, including xLAM-2-70B-FC-R~\citep{prabhakar2025apigen}, GPT-4o~\citep{hurst2024gpt}, OpenAI o1~\citep{jaech2024openai}, Qwen3-32B~\citep{yang2025qwen3}, DeepSeek-R1~\citep{guo2025deepseek}, Gemini-2.5-Pro~\citep{comanici2025gemini}, and Claude-3.7-Sonnet~\citep{anthropic2025claude}. We prepare preference pairs for each data task according to its turn-wise trajectory pattern. For single-turn tasks (splits originally introduced in BFCL v1 and v2), evaluation is based on the Abstract Syntax Tree (AST), which compares a model-generated function against its function documentation and a set of possible correct answers. In these cases, we source the oracle answers directly from the benchmark as the \textit{chosen} responses and extract incorrect responses from the failed trajectories, forming \textit{chosen–rejected} pairs for each task.

For multi-turn tasks (the split introduced in BFCL v3), evaluation instead relies on state-based and response-based checks, which differ from the rule-based matching used to check tool calls in building $\mathcal{D}_\text{pref}$\footnote{\url{https://gorilla.cs.berkeley.edu/blogs/13_bfcl_v3_multi_turn.html}}. In these complex scenarios, while pinpointing the single failing tool call is difficult, one can easily identify the entire incorrect turn by comparing the generated trajectory to the ground truth. We leverage this to create evaluation pairs: the incorrect output is the concatenation of all tool calls the model generated in that turn, and the correct output is the concatenation of all tool calls from the corresponding ground-truth solution. We show statistics of the enhanced reward model benchmark {\benchmark} in Figure~\ref{fig:benchmark_statistics}.

To ensure fair evaluation across different types of baseline models, we first apply the same \textit{think-mode}/\textit{no-think-mode} template used in our model evaluations. If the test model is unable to follow the specific instruction, we instead evaluate it using its official prompt. To fully harness the potential of the test models, the official default sampling parameters are used for inference, except that the maximum output length is limited to 8,192 tokens to prevent excessively long and repetitive chain-of-thought content.

\begin{table*}[!t]
\centering
\caption{Detailed policy model accuracy (\%) on BFCL-v3 before and after RL training with ToolRM-Gen.}
\label{tab:detail_model_rl_res}
\scalebox{0.53}{
\begin{tabular}{lcccccccccccc}
\toprule
\multirow{2}{*}{\textbf{Model}} & \multicolumn{4}{c}{\textbf{Non-Live AST}} & \multicolumn{4}{c}{\textbf{Live AST}} & \multicolumn{4}{c}{\textbf{Multi-Turn}} \\ \cline{2-13} 
 & \begin{tabular}[c]{@{}c@{}}Non-Live\\ Simple\end{tabular} & \begin{tabular}[c]{@{}c@{}}Non-Live\\ Multiple\end{tabular} & \begin{tabular}[c]{@{}c@{}}Non-Live\\ Parallel\end{tabular} & \begin{tabular}[c]{@{}c@{}}Non-Live\\ Parallel Multiple\end{tabular} & \begin{tabular}[c]{@{}c@{}}Live\\ Simple\end{tabular} & \begin{tabular}[c]{@{}c@{}}Live\\ Multiple\end{tabular} & \begin{tabular}[c]{@{}c@{}}Live\\ Parallel\end{tabular} & \begin{tabular}[c]{@{}c@{}}Live\\ Parallel Multiple\end{tabular} & \begin{tabular}[c]{@{}c@{}}Multi Turn\\ Base\end{tabular} & \begin{tabular}[c]{@{}c@{}}Multi Turn\\ Miss Func\end{tabular} & \begin{tabular}[c]{@{}c@{}}Multi Turn\\ Miss Param\end{tabular} & \begin{tabular}[c]{@{}c@{}}Multi Turn\\ Long Context\end{tabular} \\
\midrule
Qwen3-4B-Instruct-2507 & 73.83 & 94.50 & 91.00 & 88.50 & 79.84 & 76.26 & 56.25 & 75.00 & 27.00 & 10.50 & 18.00 & 24.00 \\
\rowcolor{lightgreen}
- \textit{RL w/ ToolRM-Gen} & 71.75 & 94.50 & 90.50 & 87.50 & 81.78 & 78.25 & 68.75 & 62.50 & 41.00 & 9.50 & 20.50 & 31.00 \\
\bottomrule
\end{tabular}
}
\end{table*}

\begin{figure}[!t]
\centering
\subfloat[Evaluation results on {\benchmark}.\label{fig:data_scale_performance}]{%
  \includegraphics[width=0.48\columnwidth]{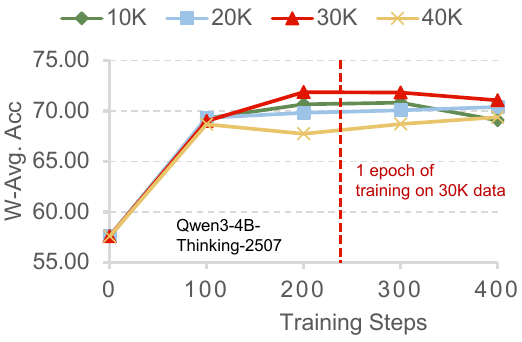}%
}\hfill
\subfloat[Statistics across different data scales.\label{fig:data_scale_statistics}]{%
  \includegraphics[width=0.48\columnwidth]{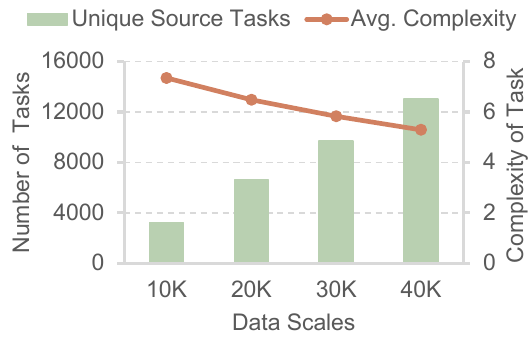}%
}
\caption{Statistics and the impact of data scale on model training.}
\label{fig:impact_of_data_scale}
\end{figure}

\subsection{Policy Model Training}
\label{app:model_rl_setup}
In the policy-model RL training phase using {\model} as the reward model, we begin by collecting 10,000 multi-turn agent tool-use trajectories spanning 2,000 unique tool schemas. For each trajectory, we randomly select one tool-call assistant response and construct a training query from all preceding messages, discarding any subsequent messages (including the selected assistant response). This yields 11,500 multi-turn queries and 3,500 single-turn queries. We then run two epochs of GRPO training with a learning rate of \texttt{2e-6} and a clip ratio of $\epsilon=0.2$. At each training step, we sample a batch of 512 queries and generate 8 trajectories per query. During policy training, we cap the maximum prompt length at 16,384 tokens and the maximum response length at 2,048 tokens.
Table~\ref{tab:detail_model_rl_res} reports detailed BFCL-v3 policy model accuracy before and after RL training with {\model}-Gen.

\section{Impact of Data Scaling on {\model}}
\label{app:data_scaling}
We investigate the influence of data scaling on model performance. Figure~\ref{fig:data_scale_performance} shows the results for Qwen3-4B-Thinking-2507 on {\benchmark}, trained with data samples ranging from 10K to 40K. Notably, the model achieves its highest performance with 30K training samples. Performance does not increase monotonically with data size because our sampling strategy prioritizes more complex tasks. As the dataset grows, the average task complexity declines, leading to less effective training signals. Figure~\ref{fig:data_scale_statistics} illustrates this trend: while the number of unique tasks rises with larger datasets, their average complexity decreases. These results demonstrate that our proposed strategy successfully balances task diversity and complexity when exploring the candidate data pool.

\begin{table}[!t]
    \centering
    \caption{Ablated evaluation results on {\benchmark}.}
    \label{tab:ablation}
    \scalebox{0.75}{
        \begin{tabular}{lcc}
            \toprule
            \textbf{Model}       & \textbf{W-Avg. Acc} \\ \midrule
            Full {\model}-Gen        & 71.87               \\
            \hdashline
            \textit{- w/o Full BMDS} & 67.24 {\small \textcolor{darkred}{(-4.63)}}       \\
            \textit{- w/o DDS} & 68.64 {\small\textcolor{darkred}{(-3.23)}} \\
            \textit{- w/o CPI} & 70.29 {\small\textcolor{darkred}{(-1.58)}} \\
            \textit{- w/o CT} & 68.89 {\small\textcolor{darkred}{(-2.98)}} \\
            \hdashline
            \textit{- w/o EC} & 68.69 {\small\textcolor{darkred}{(-3.18)}}       \\
            \bottomrule
        \end{tabular}
    }
\end{table}

\section{Ablation Studies on Preference Data Construction}
\label{app:ablation_data_construction}
To assess the contribution of our two key data construction components, we conduct an ablation study with two sets of variants. In the first set, we replace balanced multi-dimensional sampling with random sampling (\textit{w/o BMDS}) and perform fine-grained ablations along three critical dimensions: diversity of data sources (\textit{w/o DDS}), coverage of preference intensity (\textit{w/o CPI}), and complexity of tasks (\textit{w/o CT}). In the second set, we remove the unified evaluation criteria during training (\textit{w/o EC}).
Models are trained using GRPO on Qwen3-4B-Thinking-2507 with 30K pairwise preferences, keeping all other settings fixed. As shown in Table~\ref{tab:ablation}, removing either component significantly degrades performance. Each BMDS dimension contributes to performance; diversity of data sources and task complexity have larger effects than preference intensity, underscoring the importance of both diversity and contextual complexity for reward-model training. Moreover, output length of models decreases sharply without the evaluation criteria (1{,}204$\rightarrow$694), suggesting these criteria promote more comprehensive reasoning during training.

\section{Ablation Studies on Model Training}
\label{app:ablation_model_training}
\begin{table}[!t]
    \centering
    \caption{Evaluation results of different variants on {\benchmark}.}
    \label{tab:ablation_2_1}
    \scalebox{0.75}{
        \begin{tabular}{lcc}
            \toprule
            \textbf{Model} & \textbf{W-Avg. Acc} \\ \midrule
            Qwen3-4B-Instruct-2507 & 59.67               \\
            \hdashline
            - GenRM on \textit{NormalPref} &63.82 {\small \textcolor{darkgreen}{(+4.15)}} \\
            \textbf{- GenRM on \textit{ToolPref}} & \textbf{66.85 {\small \textcolor{darkgreen}{(+7.18)}}} \\
            \hdashline
            - DiscRM on \textit{NormalPref} & 67.88 {\small \textcolor{darkgreen}{(+8.21)}} \\
            \textbf{- DiscRM on \textit{ToolPref}} &\textbf{77.61 {\small \textcolor{darkgreen}{(+17.94)}}} \\
            \midrule
            Qwen3-4B-Thinking-2507 & 57.59               \\
            \hdashline
            - GenRM on \textit{NormalPref} &63.19 {\small \textcolor{darkgreen}{(+5.60)}} \\
            \textbf{- GenRM on \textit{ToolPref}} & \textbf{71.87 {\small \textcolor{darkgreen}{(+14.28)}}} \\
            \hdashline
            - DiscRM on \textit{NormalPref} & 69.69 {\small \textcolor{darkgreen}{(+12.10)}} \\
            \textbf{- DiscRM on \textit{ToolPref}} &\textbf{76.80 {\small \textcolor{darkgreen}{(+19.21)}}} \\
            \bottomrule
        \end{tabular}
    }
\end{table}
\paragraph{Data Domain.}
We investigate the influence of in-domain preference data on reward model performance by conduct the following experiments: (i) we randomly sample 30,000 instances from \textit{Skywork-Reward-Preference-80K-v0.2}~\citep{liu2024skywork}, a high-quality general preference dataset; (ii) we make minimal modifications to {\model} prompt template (removing the original evaluation criteria) and use it to perform RL training on the baseline models in the same way as for previous {\model}; (iii) the trained models are then evaluated on {\benchmark} where evaluation results are labeled with \textit{NormalPref} in Table~\ref{tab:ablation_2_1}. According to the results, models trained on high-quality normal preference data do improve their judging performance on pairwise classification tasks in the tool-use domain. However, the in-domain preference dataset delivers substantially larger gains over base models, particularly when training from a think-version base model.

\paragraph{Training Objective.}
Following \citet{liu2024skywork}, we further investigate the impact of {\dataset} on training discriminative reward models (DiscRM) using the Bradley–Terry (BT) objective. As shown in Table~\ref{tab:ablation_2_1}, the constructed dataset remains effective under BT objective and can further improve RM performance compared with the RL objective in pairwise preference classification tasks. This is consistent with our previous findings on the Skywork-Critic/Reward model series: when using the same base model and training data, DiscRM trained with a BT objective naturally produces more accurate relative scores than GenRM.
We also observe that instruct-tuned base models, which produce more concise outputs, are better suited to train DiscRM with a BT objective for generating precise scores, whereas think-version models, which produce longer initial chain of thoughts and exhibit stronger exploration capability, are better suited for RL training to obtain GenRM with stronger analytical ability. As shown in Table~\ref{tab:ablation_2_2}, {\model}-Disc yields larger gains when used to judge best-of-N sampling, whereas {\model}-Gen is substantially more effective when used to provide self-correction feedback.
In practice, each training objective has distinct strengths and should be chosen according to the application scenario: use GenRM when critique-style feedback and interpretability are required (e.g., self-correction), and use DiscRM when only accurate reward scoring is needed (e.g., RL training or BoN sampling).

\begin{table}[!t]
    \centering
    \caption{Evaluation results of different variants on ACEBench.}
    \label{tab:ablation_2_2}
    \scalebox{0.75}{
        \begin{tabular}{lcc}
            \toprule
            \textbf{Model} & \textbf{Acc} \\ \midrule
            Qwen3-4B-Instruct-2507 & 63.4               \\
            \hdashline
            - BoN-16 w/ ToolRM-Gen & 66.6 {\small \textcolor{darkgreen}{(+3.2)}} \\
            - BoN-16 w/ ToolRM-Disc & 67.2 {\small \textcolor{darkgreen}{(+3.8)}} \\
            - \textbf{SC w/ ToolRM-Gen} & \textbf{74.8 {\small \textcolor{darkgreen}{(+11.4)}}} \\
            \bottomrule
        \end{tabular}
    }
\end{table}

\section{The Balanced Multi-Dimensional Sampling Algorithm}
\label{app:algo_BMDS}
In this section, we detail the implementation of the BMDS strategy for efficient sampling. To discretize the distribution of preference intensities $I_\text{preference}$ among data samples, we initialize a set of bins $B=\{b_0,b_1,\dots, b_m\}$ with fixed intervals. In our experiments, we set: $B=\{(0, 0.1], (0.1, 0.2], \dots, (0.9, 1]\}$.
Each sample in the candidate pairwise data pool $\mathcal{D}_\text{pair-cand}$ is assigned to the corresponding bin, indexed from 0 to $m$, according to its preference intensity. We then group the samples by a composite key (\texttt{source}, \texttt{bin\_index}) to ensure representation across different data sources and varying preference intensities. Within each group, samples are sorted in descending order of task complexity $S_\text{complexity}$.
Sampling proceeds greedily: we first exhaustively select all samples from the group with the fewest entries, and then allocate the remaining quota as evenly as possible across the other bins. This yields a diverse, well-balanced, and sufficiently challenging subset of data. We present pseudocode of this strategy in Algorithm~\ref{alg:BMDS_strategy}.

\section{Example of Tool-Use Task Trajectory}
\label{app:example_trajectory}
\begin{figure*}[!t]
\centering
\includegraphics[width=\linewidth]{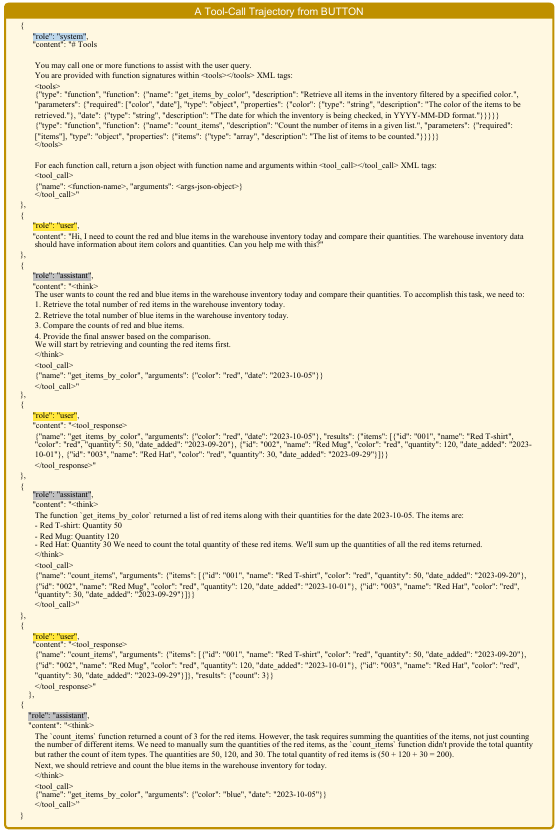}
\caption{A format-aligned tool-use trajectory from BUTTON.}
\label{fig:example_trajectory}
\end{figure*}

During conversation order validation, we retain only trajectories that satisfy the following message-role transition rules: [\texttt{system}$\rightarrow$\texttt{user}, \texttt{user}$\rightarrow$\texttt{assistant}, \texttt{assistant}$\rightarrow$\texttt{user/tool}, \texttt{tool}$\rightarrow$\texttt{assistant}]. In this work, tool responses are set into \texttt{user} messages for compatibility. Figure~\ref{fig:example_trajectory} shows a format-aligned example from BUTTON illustrating a tool-use task trajectory.

\section{Prompting Templates}
\label{app:prompt_templates}
\begin{figure*}[!t]
\centering
\includegraphics[width=\linewidth]{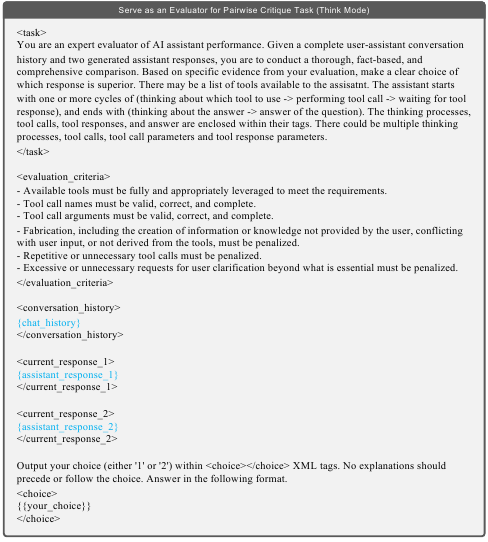}
\caption{Evaluator prompt template of the pairwise critique task for reasoning LLMs.}
\label{fig:pairwise_critique_task_think_prompt}
\end{figure*}

\begin{figure*}[!t]
\centering
\includegraphics[width=\linewidth]{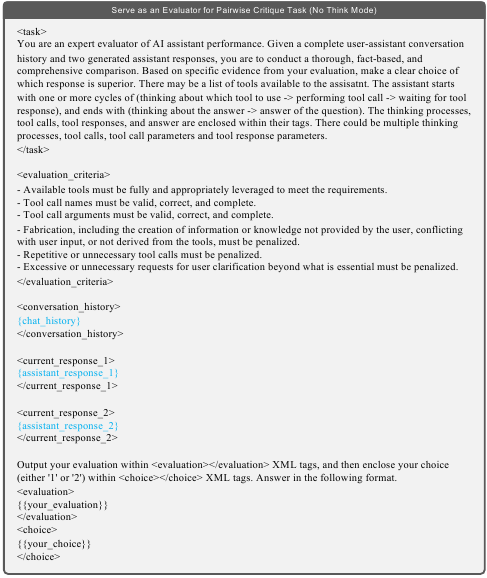}
\caption{Evaluator prompt template of the pairwise critique task for non-reasoning LLMs.}
\label{fig:pairwise_critique_task_no-think_prompt}
\end{figure*}

\begin{figure*}[!t]
\centering
\includegraphics[width=\linewidth]{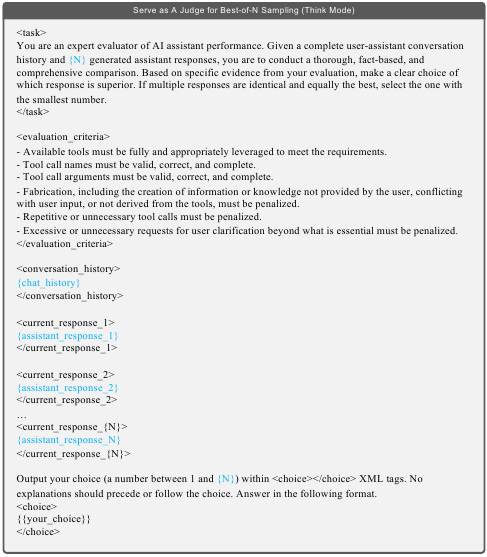}
\caption{Judge prompt template of the Best-of-N sampling task for reasoning LLMs.}
\label{fig:BoN_judge_prompt}
\end{figure*}

\begin{figure*}[!t]
\centering
\includegraphics[width=\linewidth]{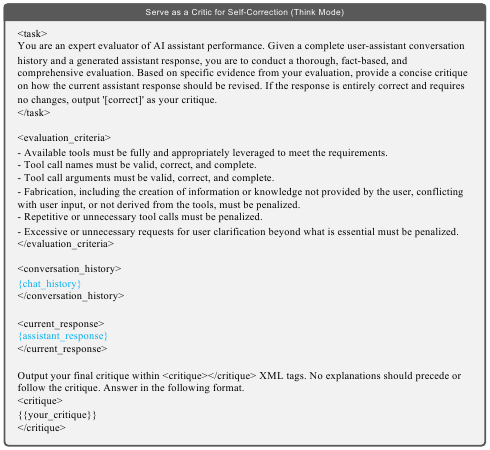}
\caption{Critic prompt template of the self-correction task for reasoning LLMs.}
\label{fig:self_correct_critic_prompt}
\end{figure*}

\begin{figure*}[!t]
\centering
\includegraphics[width=\linewidth]{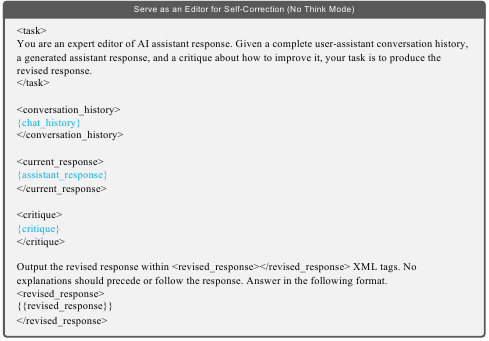}
\caption{Editor prompt template of the self-correction task for non-reasoning LLMs.}
\label{fig:self_correct_editor_prompt}
\end{figure*}

\begin{figure*}[!t]
\centering
\includegraphics[width=\linewidth]{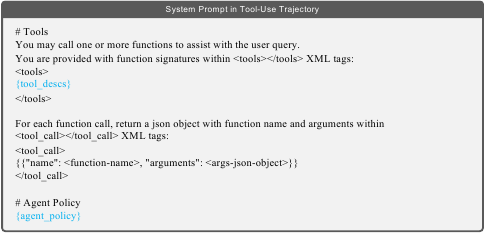}
\caption{System prompt template of the tool-use trajectory.}
\label{fig:system_prompt}
\end{figure*}

We present the evaluator prompt templates for the pairwise critique task used in both training and inference. The \textit{think-mode} and \textit{no-think-mode} templates are shown in Figures~\ref{fig:pairwise_critique_task_think_prompt} and~\ref{fig:pairwise_critique_task_no-think_prompt}, respectively. Figure~\ref{fig:BoN_judge_prompt} shows the prompt template used by the judge for the BoN sampling task, in which the $N$ sampled responses are inserted and labeled from 1 to $N$. Figures~\ref{fig:self_correct_critic_prompt} and ~\ref{fig:self_correct_editor_prompt} show the prompt templates used by critic and editor for the self-correction task. Figure~\ref{fig:system_prompt} presents the template of the system prompt in each tool-use trajectory.

\section{Case Studies}
\label{app:case_study}
\begin{figure*}[!t]
\centering
\includegraphics[width=\linewidth]{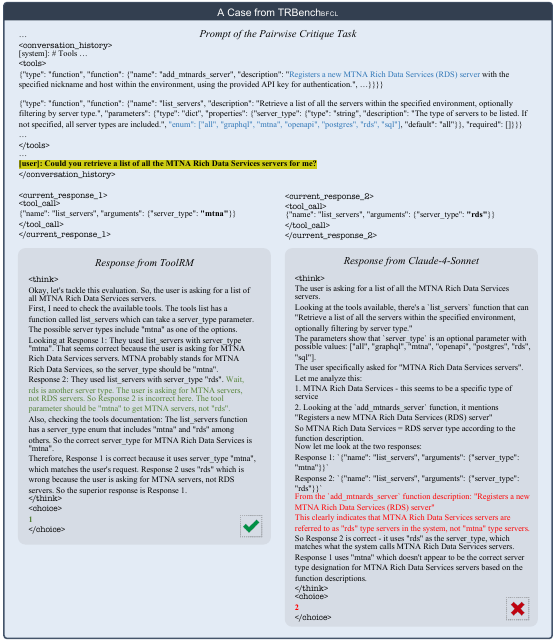}
\caption{Valid \textit{case 1} from {\benchmark}, featuring responses from {\model}-Gen and Claude 4 Sonnet.}
\label{fig:case_study_1}
\end{figure*}

\begin{figure*}[!t]
\centering
\includegraphics[width=\linewidth]{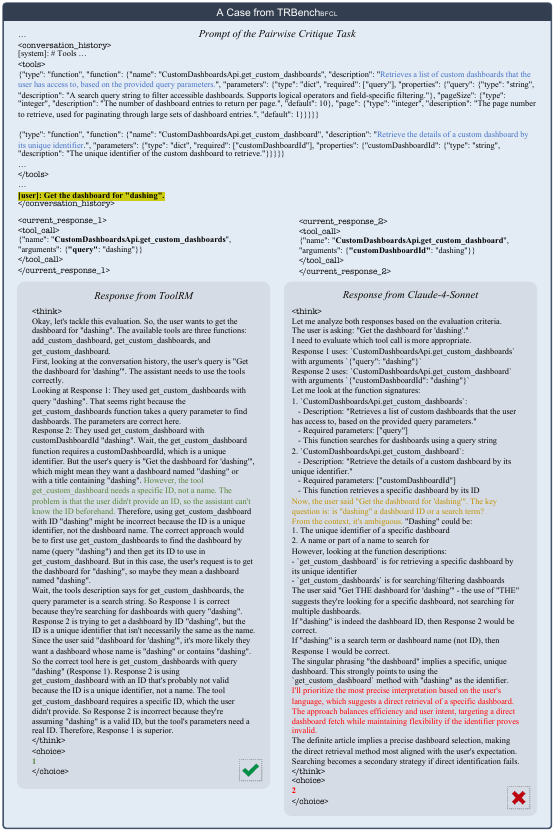}
\caption{Valid \textit{case 2} from {\benchmark}, featuring responses from {\model}-Gen and Claude 4 Sonnet.}
\label{fig:case_study_2}
\end{figure*}

\begin{figure*}[!t]
\centering
\includegraphics[width=\linewidth]{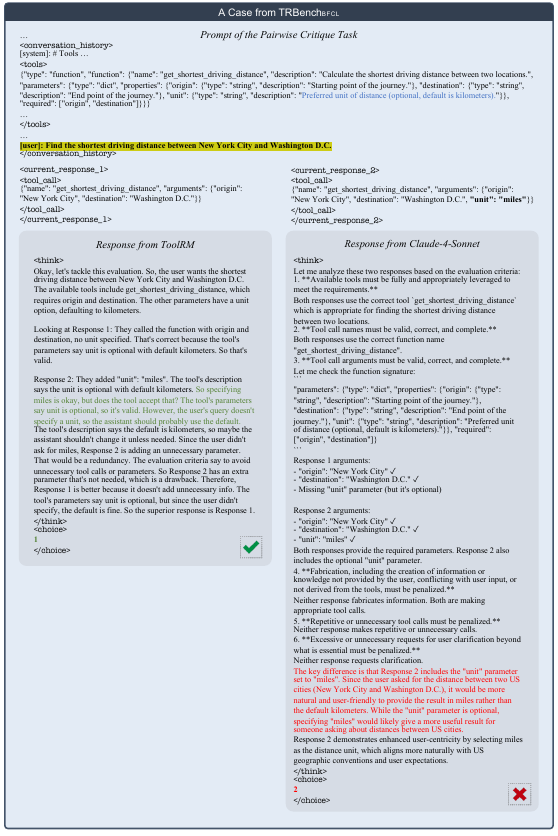}
\caption{Valid \textit{case 3} from {\benchmark}, featuring responses from {\model}-Gen and Claude 4 Sonnet.}
\label{fig:case_study_3}
\end{figure*}

\paragraph{Valid Cases}
Through representative valid cases, we compare critiques from Claude 4 Sonnet and {\model}-Gen-Qwen3-4B-Thinking-2507 on {\benchmark} test samples. In the case shown in Figure~\ref{fig:case_study_1}, {\model} accurately distinguishes correct from incorrect tool-call parameters without inducing “overthinking” hallucinations when the user query plausibly maps to multiple candidate parameters. Another case in Figure~\ref{fig:case_study_2} further demonstrates its tendency to ground analysis in contextual rationale rather than engage in speculative, divergent reasoning. Moreover, as shown in Figure~\ref{fig:case_study_3}, {\model} adheres more closely to the evaluation criteria, preferring tool calls without redundant parameters. Taken together, these behaviors enable {\model} to deliver reliable critiques in tool-use scenarios.

\begin{figure*}[!t]
\centering
\includegraphics[width=\linewidth]{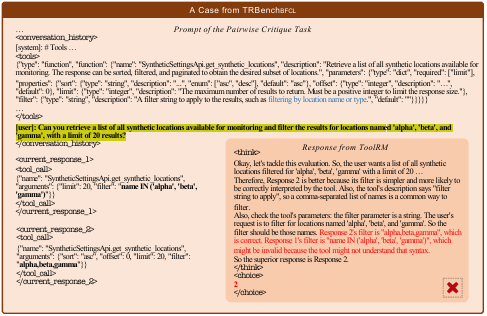}
\caption{Error \textit{case 1} from {\benchmark}, featuring responses from {\model}-Gen.}
\label{fig:error_case_1}
\end{figure*}

\begin{figure*}[!t]
\centering
\includegraphics[width=\linewidth]{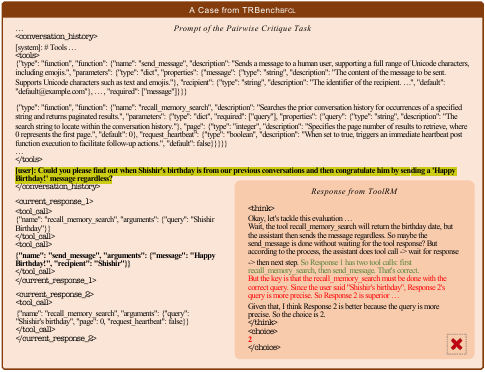}
\caption{Error \textit{case 2} from {\benchmark}, featuring responses from {\model}-Gen.}
\label{fig:error_case_2}
\end{figure*}

\paragraph{Error Cases}
We also present representative error cases of {\model}-Gen on {\benchmark}, as discussed in Section~\ref{sec:error_analysis}. Figure~\ref{fig:error_case_1} illustrates an error of type (i), where the model fails to reason correctly given an underspecified tool parameter description, while Figure~\ref{fig:error_case_2} illustrates an error of type (ii), where the model fails to distinguish primary errors from secondary ones.

\clearpage

\begin{algorithm*}[!t]
\caption{Balanced Multi-Dimensional Sampling Strategy}
\label{alg:BMDS_strategy}

\begin{algorithmic}[1]
\REQUIRE Data pool $\mathcal{D}_\text{pair-cand}$, bin edges $B$, target sample size $N$
\ENSURE A subset $\mathcal{D}_\text{pair-sampled}$ of diverse, balanced, and challenging samples

\STATE \textit{\# Step 0: Check data sufficiency}
\IF{$|\mathcal{D}_\text{pair-cand}| < N$}
    \STATE \textbf{raise} InsufficientDataError
\ENDIF

\STATE \textit{\# Step 1: Assign samples to bins}
\FOR{each $d_i \in \mathcal{D}_\text{pair-cand}$}
    \STATE $d_i.\texttt{bin\_idx} \leftarrow \text{assign}(d_i.I_\text{preference}, B)$
\ENDFOR

\STATE \textit{\# Step 2: Group by composite key}
\STATE Initialize group dictionary $\mathcal{G} \leftarrow \emptyset$
\FOR{each $d_i \in \mathcal{D}_\text{pair-cand}$}
    \STATE $key \leftarrow (d_i.\texttt{source}, d_i.\texttt{bin\_idx})$
    \STATE $\mathcal{G}[key] \leftarrow \mathcal{G}[key] \cup \{d_i\}$
\ENDFOR

\STATE \textit{\# Step 3: Sort within each group by task complexity (descending)}
\FOR{each group $G \in \mathcal{G}$}
    \STATE $G \leftarrow \text{sort}(G, \text{key}=S_\text{complexity}, \text{order=descending})$
\ENDFOR

\STATE \textit{\# Step 4: Sort groups by size (ascending)}
\STATE $\mathcal{G}_\text{sorted} \leftarrow \text{sort}(\mathcal{G}.\text{values}(), \text{key}=|G|, \text{order=ascending})$

\STATE \textit{\# Step 5: Greedy allocation}
\STATE Initialize sampling quotas: $Q \leftarrow [0] \times |\mathcal{G}_\text{sorted}|$
\STATE $N_\text{remaining} \leftarrow N$, $k \leftarrow 0$

\WHILE{$k < |\mathcal{G}_\text{sorted}|$ \AND $N_\text{remaining} > 0$}
    \STATE $m \leftarrow |\mathcal{G}_\text{sorted}| - k$
    \STATE $n_\text{avg} \leftarrow \lceil N_\text{remaining} / m \rceil$

    \IF{$|\mathcal{G}_\text{sorted}[k]| \leq n_\text{avg}$}
        \STATE $Q[k] \leftarrow |\mathcal{G}_\text{sorted}[k]|$
        \STATE $N_\text{remaining} \leftarrow N_\text{remaining} - |\mathcal{G}_\text{sorted}[k]|$
        \STATE $k \leftarrow k + 1$
    \ELSE
        \STATE \textit{\# Distribute remaining quota evenly}
        \STATE $q \leftarrow \lfloor N_\text{remaining} / m \rfloor$
        \STATE $r \leftarrow N_\text{remaining} \mod m$
        \FOR{$i = k$ \TO $|\mathcal{G}_\text{sorted}| - 1$}
            \STATE $Q[i] \leftarrow q$
        \ENDFOR
        \FOR{$i = 0$ \TO $r - 1$}
            \STATE $Q[|\mathcal{G}_\text{sorted}| - 1 - i] \leftarrow Q[|\mathcal{G}_\text{sorted}| - 1 - i] + 1$
        \ENDFOR
        \STATE \textbf{break}
    \ENDIF
\ENDWHILE

\STATE \textit{\# Step 6: Sample data based on quotas}
\STATE $\mathcal{D}_\text{pair-sampled} \leftarrow \emptyset$
\FOR{$i = 0$ \TO $|\mathcal{G}_\text{sorted}| - 1$}
    \STATE $\mathcal{D}_\text{pair-sampled} \leftarrow \mathcal{D}_\text{pair-sampled} \cup \mathcal{G}_\text{sorted}[i][:Q[i]]$
\ENDFOR

\RETURN $\mathcal{D}_\text{pair-sampled}$
\end{algorithmic}
\end{algorithm*}

\clearpage

\clearpage

\clearpage

\clearpage

\clearpage

\clearpage

\clearpage

\clearpage

\end{document}